%% file: 0-paper-qa.tex
\documentclass[11pt]{article}

\usepackage[preprint]{acl}

\usepackage{times}
\usepackage{latexsym}

\usepackage[T1]{fontenc}

\usepackage[utf8]{inputenc}

\usepackage{microtype}

\usepackage{inconsolata}

\usepackage{graphicx}

\usepackage{hyperref}
\usepackage{url}
\usepackage{xspace}
\usepackage{enumitem}
\usepackage{amsmath}
\usepackage{listings}

\usepackage{array}
\usepackage{booktabs}
\usepackage{multirow}

\usepackage{subcaption}

\usepackage{xcolor}  
\usepackage{pifont}  
\usepackage{stackengine}
\usepackage[most]{tcolorbox}
\usepackage{wrapfig}
\usepackage{amsfonts}  

\usepackage{titlesec}

\titlespacing*{\section}
{0pt}          
{12pt}         
{6pt}          

\newcommand{\benchmark}{RPC-Bench\xspace}

\lstdefinestyle{prompt}{
    basicstyle=\ttfamily\small,
    backgroundcolor=\color{gray!10},
    frame=single,
    breaklines=true,
    columns=fullflexible,
    keepspaces=true,
    showstringspaces=false
}

\newcounter{examplectr}

\newenvironment{exampleblock}[1][]%
{%
  \refstepcounter{examplectr}
  \par\medskip
  \noindent\textbf{Example~\theexamplectr%
  \if\relax\detokenize{#1}\relax%
  \else\ (#1)\fi:}%
  \par\vspace{0.5\baselineskip}
  \small
  \setlength{\parindent}{0pt}   
}%
{%
  \par\medskip
  \normalsize
}

\title{RPC-Bench: A Fine-grained Benchmark for Research Paper Comprehension}

\author{
 \textbf{Yelin Chen\textsuperscript{1}}\thanks{Equal contribution. Email: \texttt{yelin@stu.xju.edu.cn}}\thanks{Part of the work was done when Yelin interned at Z.ai.},
 \textbf{Fanjin Zhang\textsuperscript{2}}\footnotemark[1]\thanks{Corresponding authors. Email: \texttt{fanjinz@ruc.edu.cn}, \texttt{lijuanzi@mail.tsinghua.edu.cn}. Fanjin Zhang is with School of Information, Renmin University of China, Key Laboratory of Data Engineering and Knowledge Engineering, and Engineering Research Center of Database and Business Intelligence.},
 \textbf{Suping Sun\textsuperscript{3}},
 \textbf{Yunhe Pang\textsuperscript{4}},
 \textbf{Yuanchun Wang\textsuperscript{2}},
\\
 \textbf{Jian Song\textsuperscript{5}},
 \textbf{Xiaoyan Li\textsuperscript{5}},
 \textbf{Lei Hou\textsuperscript{6}},
 \textbf{Shu Zhao\textsuperscript{3}},
 \textbf{Jie Tang\textsuperscript{6,}\textsuperscript{7}},
 \textbf{Juanzi Li\textsuperscript{6}}\footnotemark[3]
\\
 \textsuperscript{1}Xinjiang University,
 \textsuperscript{2}Renmin University of China,
 \textsuperscript{3}Anhui University,
\\
 \textsuperscript{4}Sun Yat-sen University,
 \textsuperscript{5}Z.ai,
 \textsuperscript{6}Tsinghua University,
 \textsuperscript{7}University of Southampton
}

\newcommand{\hide}[1]{}

\begin{document}
\maketitle
\begin{abstract}
Understanding research papers remains challenging for foundation models due to specialized scientific discourse and complex figures and tables, yet existing benchmarks offer limited fine-grained evaluation at scale.
To address this gap, 
we introduce \benchmark, a large-scale question-answering benchmark built from review–rebuttal exchanges of high-quality computer science papers, containing 15K human-verified QA pairs.
We design a fine-grained taxonomy aligned with the scientific research flow to assess models’ ability to understand and answer why, what, and how questions in scholarly contexts.
We also define an elaborate LLM–human interaction annotation framework to support large-scale labeling and quality control.
Following the LLM-as-a-Judge paradigm, we develop a scalable framework that evaluates models on correctness-completeness and conciseness, with high agreement to human judgment.
Experiments reveal that even the strongest models (GPT-5) achieve only 68.2\% correctness-completeness, dropping to 37.46\% after conciseness adjustment, highlighting substantial gaps in precise academic paper understanding.
Our code and data are available\footnote{\url{https://rpc-bench.github.io/}}.
\end{abstract}

\input{1-Introduction}

\input{2-RelatedWork}

\input{3-Benchmark}

\input{4-Experiments}

\input{5-Conclusion}

\section*{Limitations}

The current benchmark provides substantial coverage and diversity within its defined scope, yet several aspects remain open for further exploration.

Due to the high cost of manual annotation, we employed multiple LLMs to reformulate review–rebuttal pairs into question–answer form without altering their original information. Manual annotation was prioritized for the development and test sets to ensure robust evaluation quality, whereas the training portion retains LLMs‑reformulated QA pairs for researcher‑specified selection, processing, and usage.

While \benchmark is collected from the high‑quality OpenReview platform and its construction process ensures topical diversity, thereby mitigating potential bias, the current release primarily covers computer science and its subfields. 
This focused scope is a natural consequence of the domain distribution of the source data and enables comprehensive evaluation within the target domain, while the established data pipeline and evaluation methodology provide a solid foundation for future expansion into additional areas such as the life sciences and social sciences.

\benchmark is specifically designed to rigorously assess model comprehension of scholarly articles, with the current stage focusing on single‑article understanding. Building on this core capability, future work will extend the evaluation to cross‑document reasoning and multi‑paper synthesis, broadening the benchmark’s applicability to more complex forms of scholarly interaction.

\section*{Ethical Considerations}

This work adheres to the ACL Code of Ethics. The benchmark introduced in this study (\benchmark) is constructed exclusively from publicly available academic papers and their associated review–rebuttal pairs hosted on OpenReview. All source materials were originally authored for public scholarly dissemination, and no private, confidential, or proprietary information is included. Our use of OpenReview content is limited to non-commercial academic research and is consistent with its terms of service and copyright policies; we do not claim ownership of the original texts.

All released artifacts, including the derived question–answer annotations and annotation guidelines, will be distributed under the Creative Commons Attribution 4.0 International (CC BY 4.0) license, with appropriate attribution to the original authors and platform. The derived dataset is intended solely for research purposes such as benchmarking, analysis, and methodological development, and must not be used for production or commercial deployment. Dataset release will fully comply with the original access conditions of the source data.

To support safe and responsible use, we anonymize the data by removing author names, reviewer identifiers, and residual metadata that could enable re-identification, making identification infeasible without significant effort. We perform quality control to remove low-quality or irrelevant content and apply balanced sampling to mitigate systematic biases in data sources. The benchmark does not contain harmful, discriminatory, or security-sensitive content. 

No experiments in this work involve personal health data or sensitive demographic attributes. Potential conflicts of interest, including affiliations or sponsorships, have been disclosed in accordance with conference policies. 
This study is intended to advance model evaluation for academic paper comprehension and is not designed to automate peer-review decisions or replace human reviewers. Any practical use of AI in review-related settings should follow strict confidentiality, data protection, and human-in-the-loop principles, with appropriate safeguards to reduce privacy and security risks.

\section*{Acknowledgments}

This work is supported by 
the National Natural Science Foundation of China (62406164, 62476003, 62425601), 
Beijing Natural Science Foundation (L243006), 
the New Cornerstone Science Foundation through the XPLORER PRIZE,
and Anhui Province Excellent Scientific Research and Innovation Team (2024AH010004). 

\bibliography{custom}

\appendix

\input{6-Appendix}

\end{document}

%% file: 1-Introduction.tex
\section{Introduction}

Large foundation models are increasingly serving as research copilots, supporting knowledge extraction~\citep{zhang2019oag,zhang2024oag,chen2025small}, deep research \citep{schmidgall2025agentlaboratoryusingllm},
and even end-to-end research automation \citep{aiscientist_v2, gottweis2025towards}. 
A key prerequisite for these applications is the ability of large foundation models to understand research papers
—not only by parsing explicit content, but also by grasping specialized concepts, analyzing methodological motivations, and evaluating experimental limitations to inform subsequent scientific discovery.

\begin{table*}[h]
\small
    \newcolumntype{C}{>{\centering\arraybackslash}c}
    \centering
    \renewcommand\arraystretch{1.0}
    \begin{tabular}{c|@{~ }*{1}{CCCCCCC}}
            \toprule[1.2pt]
        {Benchmarks}  &  {Papers}  & {QA} & {Real QA} & {Taxonomy} & {Eval. Metrics} & {Textual inp.} & {Visual inp.} \\ 
        \midrule
        PeerQA & 208   & 579 & \textcolor{green}{\ding{51}} & task & Corr. & \textcolor{green}{\ding{51}} & \textcolor{red}{\ding{55}}  \\
        SPIQA & 25.5K & 270K & \textcolor{red}{\ding{55}} & task & LLMLogScore & \textcolor{green}{\ding{51}} & \textcolor{green}{\ding{51}} \\ 
        ArXivQA & 16.6K  &- & \textcolor{red}{\ding{55}} & task & - & \textcolor{red}{\ding{55}} & \textcolor{green}{\ding{51}}  \\
        DocGenome & 500K  &- & \textcolor{red}{\ding{55}} & task & GPT-acc & \textcolor{red}{\ding{55}} & \textcolor{green}{\ding{51}}  \\
        RPC-Bench & 4150 & 61.3K & \textcolor{green}{\ding{51}} & content & Conc., F1-like & \textcolor{green}{\ding{51}} & \textcolor{green}{\ding{51}}  \\
        \bottomrule[1.2pt]
    \end{tabular}
    \caption{Comparison with relevant research paper Benchmarks. Conc.=Conciseness; Corr.=Correctness; F1-like is defined as the harmonic mean of correctness and completeness; inp.=input. ``Eval. Metrics'' are LLM-based metrics.}
    \label{table:comparison_bench}
\end{table*}

Although document understanding has advanced substantially in recent years, existing benchmarks remain insufficient for rigorously evaluating this progress.
As shown in Table \ref{table:comparison_bench}, PeerQA~\citep{baumgartner2025peerqa} is limited in scale, covering only a small number of question–answering (QA) pairs. SPIQA~\citep{pramanick2024spiqa}, DocGenome~\citep{xia2024docgenome}, and ArXivQA~\citep{li2024multimodal} rely heavily on synthetic QA pairs rather than authentic scholarly interactions. 
More broadly, these benchmarks are constrained by coarse, task-centric taxonomies, lack stratification by depth of understanding, rely on limited evaluation metrics, and often fail to jointly accommodate both textual and visual inputs.
As a result, there is still no comprehensive benchmark for evaluating deep understanding of large-scale research papers.

To address this gap, \textbf{we introduce \benchmark, a large-scale benchmark for in-depth research paper comprehension.} \benchmark is built from high-quality publications (2013–2024) on OpenReview\footnote{\url{https://openreview.net/}} and their associated review–rebuttal exchanges. Unlike synthetic datasets, our QA pairs are derived from authentic peer-review interactions and converted into question–answer format through a collaborative LLM–human workflow, ensuring that all answers are grounded in the source papers. After rigorous filtering, the final benchmark encompasses 4,150 papers and 61.3K QA pairs.

To systematically capture core aspects of paper understanding, we decompose the research workflow into a fine-grained taxonomy with 4 primary dimensions—\textbf{Concepts, Methods, Experiments, and Claim Verification}—further divided into nine categories. This taxonomy objectively reflects comprehension of a paper’s conceptual, methodological, and experimental components, and guides annotation and evaluation for nuanced assessment of research paper understanding.

In addition, we design a scalable LLM-based evaluation framework  aligned with human judgment, supporting both pure-text and rendered-page inputs to benchmark large language models (LLMs) and vision language models (VLMs). Model outputs are jointly assessed for correctness (accuracy of generated responses, akin to precision), completeness (coverage of essential content, akin to recall), and conciseness, with multiple pilot-tested LLM judges aggregated to produce stable, human-consistent scores.

We conduct extensive experiments across 28 state-of-the-art models, including 11 LLMs, 3 Document-Centric Models (DCMs), 9 VLMs, and 5 retrieval-augmented generation (RAG) models. The results show that no model fully comprehends research papers. Even the best model, GPT-5, achieved only 68.2\% on F1-like (harmonic mean of correctness and completeness), dropping to 37.46\% under the conciseness-constrained F1-like. Furthermore, for multimodal-capable LLMs, replacing text inputs with page-image inputs consistently reduced F1-like by 4.74–36.1\%, highlighting persistent weaknesses in visual reasoning over scholarly documents.
In summary, our contributions are:

\begin{itemize}[leftmargin=*]
    \item We introduce \benchmark, a \textit{large-scale} benchmark grounded in authentic review–rebuttal exchanges, featuring a \textit{fine-grained} taxonomy aligned with the \textit{research workflow} for systematic evaluation of research paper comprehension.
    \item We introduce an LLM–human collaborative annotation framework that supports large-scale QA transformation and rigorous quality control.
    \item We develop an evaluation framework that jointly assesses correctness, completeness, and conciseness, with strong alignment to human judgment.
    \item We conduct a comprehensive study of 28 advanced models, revealing fundamental limitations in both text-based and multimodal research paper understanding.
\end{itemize}

%% file: 2-RelatedWork.tex
\section{Related Work}
	
\paragraph{Methodologies for Document Question Answering.}
Document QA methodologies center on three complementary pillars: (i) large foundation models, (ii) document-centric architectures, and (iii) RAG-based approaches. 
Large foundation models
span proprietary models like GPT-5~\citep{leon2025gpt}, Claude 4.5, and Gemini 3~\citep{comanici2025gemini}, and open-source families such as the Qwen~\citep{yang2025qwen3technicalreport}, GLM~\citep{glm2024chatglmfamilylargelanguage}, and DeepSeek~\citep{liu2025deepseek} series. 

Document-centric architectures are introduced to address the structure and layout of long documents.
One line of the work (e.g., Monkey-Chat-7B and DocOwl2-8B \citep{li2024monkeyimageresolutiontext,hu2024mplugdocowl2highresolutioncompressingocrfree}) enables direct, OCR-free  understanding, avoiding error propagation from external OCR. 
Another line, exemplified by layout-aware models like \textbf{DocLLM} \cite{wang2023docllm}  and \textbf {Docopilot} \citep{,duan2025docopilot}, explicitly encodes 2D page layout to better parse complex structures like tables and forms.

RAG-based approaches mitigate models’ limited parametric knowledge on large corpora by grounding generation in retrieved evidence. 
Textual RAG methods include \textbf{RAPTOR} \citep{sarthi2024raptor}, which uses recursive clustering, as well as \textbf{HippoRAG} \citep{gutierrez2025rag} and \textbf{MemoRAG} \citep{qian2025memorag}, which optimize indexing and memory.
More broadly, 
\textbf{VisRAG} \citep{yu2024visrag} and \textbf{VDocRAG} \citep{tanaka2025vdocrag} extend retrieval to visual content, enabling evidence discovery within figures and tables. 
The RAG ecosystem is further supported by toolkits like \textbf{FlashRAG} \citep{jin2025flashrag} and explorations into alternative data structures such as knowledge graphs with \textbf{GraphRAG} \citep{edge2025localglobalgraphrag}.

\begin{figure*}[t]
\centering
\includegraphics[width=0.95\linewidth]{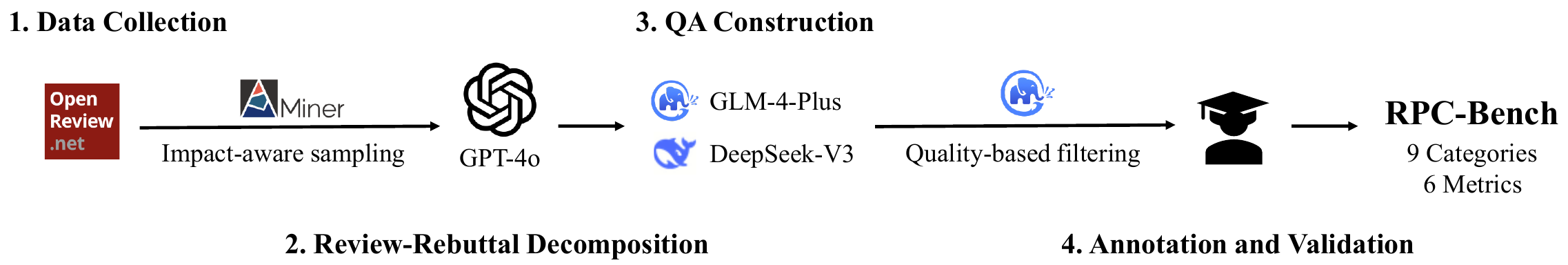}
\caption{\benchmark Construction Pipeline.
We crawl papers and review–rebuttal pairs from OpenReview and apply impact‑aware sampling to balance quality and mitigate bias.
Review-rebuttals are segmented into comment–response units with GPT‑4o, rewritten into QA pairs using GLM‑4‑Plus and DeepSeek‑V3.
Low‑quality QA items are discarded before iterative human annotation and review.
}
\label{fig:benchmark_pipeline}
\end{figure*}

\paragraph{Document QA Benchmarks.}
Numerous benchmarks have been developed to standardize the evaluation of document QA. For instance, SPIQA~\citep{pramanick2024spiqa} targets multimodal questions over figures and tables in scientific papers.
DocGenome~\citep{xia2024docgenome} offers a large-scale, multi-domain dataset for both pre-training and high-level evaluation. 
LongDocURL~\citep{deng2024longdocurl} marks a step toward finer granularity by assessing the distinct skills of understanding, reasoning, and locating.
PeerQA~\citep{baumgartner2025peerqa}, akin to our work, is text-only and relatively small in scale. It is built from reviewer-raised questions explicitly marked by question marks and therefore mainly evaluates localized clarification within peer review rather than broader paper-level comprehension.
However, existing benchmarks suffer from several notable limitations. 
(1) \textbf{Limited scale}: 
Existing benchmarks are often small in scope, failing to cover large collections of papers and QA pairs.
(2) \textbf{Quality issues}:
Many rely on automatically generated QA pairs with uncertain correctness, and they usually classify tasks only by task type rather than by the content depth.
(3) \textbf{Narrow and shallow evaluation}:
Current benchmarks tend to emphasize on multimodal QA without systematic tests of paper understanding. Moreover, they typically focus on single metrics (e.g., accuracy) that miss long-form answer quality.

Unlike prior works, 
our benchmark delivers large-scale, realistic, and accurate academic QA pairs grounded in peer reviews and rebuttals.
We categorize questions according to research stages and assess a broad spectrum of document QA methods. 
Furthermore, we introduce a scalable evaluation pipeline with high agreement with human expert evaluations.
Our comprehensive evaluation reveals 
persistent gaps in expert-level comprehension of scholarly literature.
Further comparison with other paper-related benchmarks is provided in Appendix \ref{sec:paper-related_benchmarks}.

%% file: 3-Benchmark.tex
\section{\benchmark} \label{sec:benchmark}

\begin{figure}
        \centering
		\includegraphics[width=0.75\linewidth]{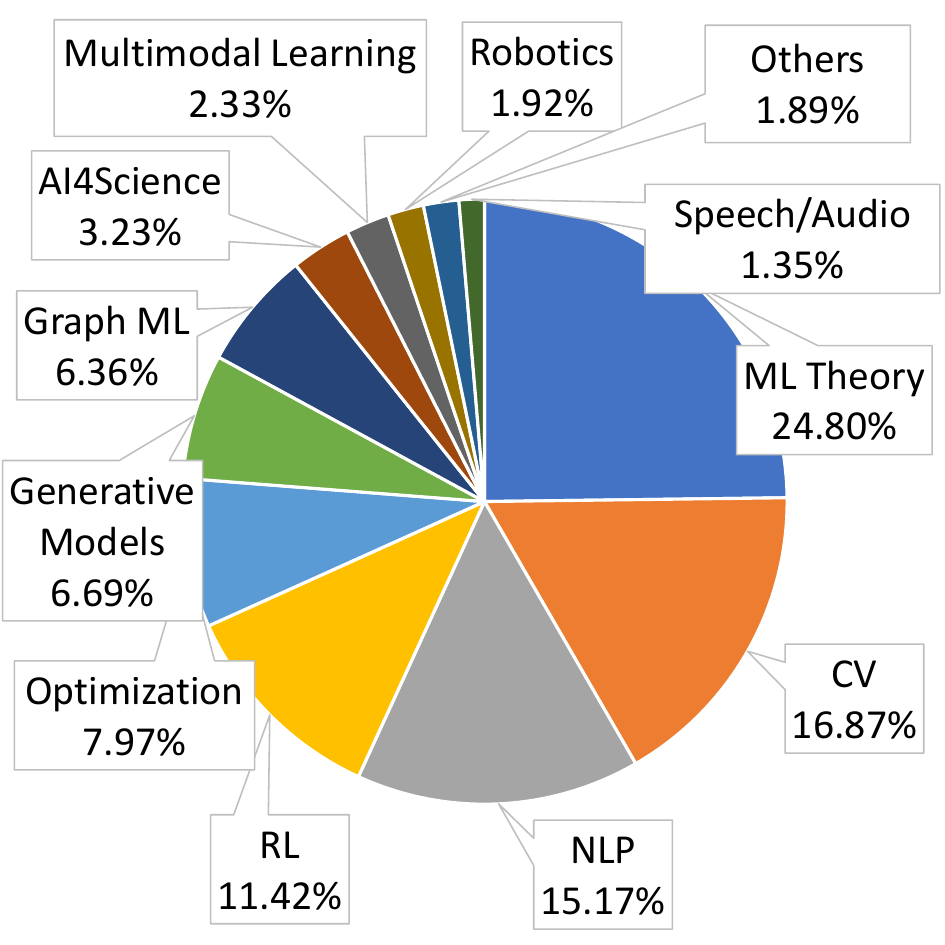}
		\caption{Domain distribution of \benchmark. ML: Machine Learning; CV: Computer Vision; NLP: Natural Language Processing; RL: Reinforcement Learning.}
        \label{fig:topic_dist}
\end{figure}

\benchmark is designed to evaluate in-depth paper comprehension under realistic  settings, emphasizing faithful understanding of concepts, methods, and experiments.
We follow a principled framework that grounds benchmark construction in authentic review–rebuttal exchanges and organizes questions according to the natural research workflow, enabling fine-grained assessment across what, how, and why dimensions. 
Figure \ref{fig:benchmark_pipeline} presents the overall framework for benchmark construction.

\begin{figure*}[t]
\centering
\begin{tcolorbox}[
    colback=blue!5!white,
    colframe=blue!50!black,
    title=Task Taxonomy,
    sharp corners,
    boxrule=0.8pt,
    before skip=0pt,
    after skip=0pt,
    parbox=false,
    width=\textwidth,
    halign title=left
]
\small
\setlength{\parskip}{0pt}
\begin{enumerate}[leftmargin=*]
    \item Concept Understanding (C.U.) [\texttt{What-4.27\%}]: Clarifies or explains key concepts, terminology, theoretical viewpoints, or information conveyed in figures, tables, or formulas.
    \item Methods
        \begin{enumerate}[label*=\arabic*.]
            \item Method Disambiguation (M.D.) [\texttt{What-8.91\%}]: Clarifies methodological details to resolve misunderstandings or ambiguities, ensuring an accurate understanding of proposed approaches.
            \item Method Mechanics (M.M.) [\texttt{How-9.91\%}]: Questions about the implementation or function of methodological workflow or components, such as the effect of specific modules in models.
            \item Motivation Analysis (M.A.) [\texttt{Why-6.29\%}]: Explores the rationale behind a proposed method or decision.
            \item Method Comparison (M.C.) [\texttt{11.75\%}]: Compares the proposed approach with baseline methods, analyzing similarities, differences, or performance to highlight novelty.
        \end{enumerate}
    \item Experiments
        \begin{enumerate}[label*=\arabic*.]
            \item Experimental Exposition (E.E.) [\texttt{What-13.07\%}]: Explains results, infers how changes to experiments could affect results or conclusions, and handling reasoning tasks (i.e., calculations, counting, or comparisons).
            \item Experimental Setup (E.S.) [\texttt{How-6.77\%}]: About the design, configuration, and execution of experiments.
            \item Experimental Analysis (E.A.) [\texttt{Why-14.08\%}]: Studies the reasons of specific experimental results, links them to the proposed approach, and assesses their generalizability and potential impact.
        \end{enumerate}
    \item Claim Verification (C.V.) [\texttt{24.95\%}]: Binary classification tasks that judge whether claims, hypotheses, or experimental conclusions are correct.
\end{enumerate}
\end{tcolorbox}
\vspace{0.2cm}
\caption{Task taxonomy of QA pairs. The form of [\texttt{What-4.27\%}] indicates question types and QA percentage.}
\label{fig:taxonomy}
\end{figure*}

\subsection{Data Collection}
To rigorously assess model capabilities in paper understanding, we built a three-stage data pipeline:

\begin{itemize}[leftmargin=*]
    \item Broad coverage collection: 
    Collected 44.7K peer‑reviewed papers with review–rebuttal pairs from OpenReview\footnote{https://openreview.net/}  (2013–2024).
    \item Quality refinement: Matched with the academic search system AMiner\footnote{https://www.aminer.cn/}~\cite{tang2008arnetminer} to remove incomplete entries, yielding a curated set of 17.7K papers.
    \item Impact‑aware sampling: 
    Selected 3521 accepted papers ($\geq 50$ citations) as positive samples, plus 361 highly-cited rejected papers and 361 random rejected papers as challenging negatives to balance quality and bias reduction.

\end{itemize}

This pipeline yields a scholarly collection of 4243 papers. We chronologically split this collection as follows: 3153 papers for training, 890 papers for validation, and 200 papers for testing. 
Note that we don't apply any topic‑based filtering to the papers. 
We use GLM-4.6 to analyze the distribution of topics. 
As shown in the Figure \ref{fig:topic_dist}, since most public venues on OpenReview focus on AI, the papers span a broad range of AI subfields.

\subsection{Taxonomy Design}

We aim to evaluate models’ understanding by probing how well they grasp the concepts, methods, experiments, and reasoning presented in scholarly articles.
To this end, we design a taxonomy aligned with the natural research flow of academic papers~\citep{sollaci2004introduction, booth2009craft}. 
It begins with \textbf{\textit{what-questions}}, which focus on clarifying fundamental concepts  and contextual background. It then advances to \textbf{\textit{how-questions}}, which probe the mechanics of methods and experimental setups. Finally, it deepens into \textbf{\textit{why-questions}}, which examine the underlying motivations of methods and the reasoning behind results. 
By moving from basic concepts to methods and then to underlying reasoning, this taxonomy helps trace the full logic of a paper, making it easier to spot concrete gaps and opportunities for future research.

Based on this principle, we define a four-level taxonomy organized around key components of research papers 
(see Figure \ref{fig:taxonomy}),
enabling fine-grained and multi-perspective coverage of academic paper understanding. Most categories are formulated as free-form QA tasks, while the Verification category is defined as a binary classification task. Both formats require models to locate, integrate, and reason over information drawn from the target paper.

To ensure our QA pairs are as objective as possible, we design the taxonomy around factual question types—what, how, why, and claim‑verification—to avoid subjective or speculative inquiry. All reference answers come directly from the original paper authors, providing an authoritative source. In addition, every answer is verified to be grounded in the camera‑ready paper, ensuring that the required information is explicitly available in the text.

\subsection{Annotation Process}\label{sec:annotation}

Manual annotation of taxonomy-based QA pairs requires domain expertise and extensive time for labeling and verification, making large-scale, high-quality data collection prohibitively costly. To mitigate this, we propose a semi-automated hybrid pipeline that leverages multiple LLMs to reduce human effort while maintaining annotation quality.

Since crawled review–rebuttal pairs from OpenReview usually contain overall reviews and general replies rather than paired comment–response matches, we first use GPT-4o to decompose each review into minimal, self-contained comment–response pairs. 
Guided by our taxonomy, GLM-4-Plus and DeepSeek-V3 are used to rewrite these pairs into free-form QA or claim verification tasks and assign each to the proper taxonomy category. 
A pilot study shows that this pipeline delivers competitive rewriting quality at a fraction of GPT‑4o’s cost, enabling scalable data generation.

To ensure the quality of the automatically generated questions, we apply a filtering process with GLM‑4‑Plus.
This process removes low‑quality items that cannot be answered from the paper itself, including \textbf{temporary or editorial issues} (e.g., grammar errors), \textbf{dependence on external resources} (e.g., external URLs or external papers), and \textbf{non‑substantive commitments} (e.g., just promise without a real answer). Additional criteria and examples are listed in Appendix \ref{sec:data_filtering}.

We employ four annotators (Master’s degree or higher), with two handling annotation and two reviewing. Before formal annotation, all annotators underwent training and practiced QA conversion on 10 sample papers, receiving iterative feedback until achieving a $\geq$95\% pass rate. To prioritize quality over speed, annotators were limited to 80 QA pairs per day, averaging 5–6 minutes per question. 
Annotated data were reviewed promptly, and problematic cases were returned for correction. 
Additional annotation details are provided in Appendix \ref{sec:detail_annotation}. 

Due to cost constraints, only the validation and test sets are manually annotated, while the training set retains QA pairs generated by LLMs. Table \ref{table:data_statistics} reports the dataset statistics.
Benchmark scale and scalability are further discussed in Appendix \ref{sec:scale_scalability}.
Refer to Appendix \ref{sec:domain_bias} for domain bias and distribution analysis, and to Appendix \ref{sec:bias_circularity} for potential bias and circularity from LLM use. 

\begin{table}[h]
    \newcolumntype{C}{>{\centering\arraybackslash}c}
    \centering
    \renewcommand\arraystretch{1.0}
    \begin{tabular}{c@{~ }*{3}{C}}
            \toprule[1.2pt]
        {Statistics} & {train} & {val} & {test}  \\ 
        \midrule
        Papers & 3100 & 850 & 200  \\
        Accept & 1980 & 609 & 116 \\
        year & 2013-2021 & 2022-2023 & 2024 \\
        Venue & 12 & 7 & 4 \\
        QA & 45651 & 12895 & 2787  \\ 
        A/M Q & 25.2/105 & 26.2/261 & 24.2/250  \\
        A/M A & 70.5/297 & 121.5/1337 & 87.9/773  \\
        \bottomrule[1.2pt]
    \end{tabular}
    \captionof{table}{Statistics of the \benchmark.
    A/M Q: average/max question length.
    A/M A: average/max answer length.
    Lengths are measured in words.
    } 
    \label{table:data_statistics}
\end{table}

\subsection{Quality Control} \label{sec:quality_control} 

To ensure high data quality and restrict all questions to be answerable solely from the source paper, we design a quality control process as follows.
First, we collect the authors’ final camera‑ready papers, which include clarifications, additional experiments, and supplementary content, 
ensuring that all answer‑relevant information is in the paper.

During annotation, reviewers check the annotated data, return problematic cases to annotators for correction, and jointly validate question answerability. This stage further remove 8.87\% of QA items deemed low-quality or unanswerable based on the corresponding paper alone. For QA items referencing specific numbers, formulas, bibtex, or section indices, annotators and reviewers are required to verify their presence in the final paper version and update all indices accordingly to maintain positional accuracy and consistency. 

We score labeling agreement via Cohen’s Kappa. For category assignment, agreement reaches 0.72 among annotators and 0.78 among reviewers. Regarding question retention, the scores are 0.81 and 0.85, respectively. These values imply strong consensus in both task understanding and judgment.
We also discuss potential data contamination from publicly available papers and its implications for benchmark reliability in Appendix \ref{sec:data_contamination}.

\begin{table*}[t]
\small
    \newcolumntype{C}{>{\centering\arraybackslash}p{3.2em}}
    \centering
    \renewcommand\arraystretch{1.0}
    \begin{tabular}{c|c|@{~ }*{1}{CC|}*{1}{CCCCC}}
            \toprule[1.2pt]
        \multirow{2}{*}{\shortstack{Model\\Type}}    
            &\multirow{2}{*}{Model } 
            &\multicolumn{2}{c|}{Traditional}
            &\multicolumn{5}{c}{LLM-as-a-judge}
        \\
        \cmidrule{3-4} \cmidrule{5-9}
        & & {R-L} & {BERTS.} & {Concise.} & {Correct.} & {Complete.} & \mbox{F1-like} & {Info.} \\
        \midrule
        \multirow{6}{*}{LLM}
        & DeepSeek-V3.2 & 19.22 & \underline{55.60} & 56.31 & 58.73 & 55.19 & 56.91 & 32.04  \\
        & GLM-4.7 & 17.09 & 48.58 & 54.34 & 54.36 & 51.75 & 53.02 & 28.81  \\
        & Qwen3 & 16.16 & 54.25 & 41.44 & 55.88 & 56.64 & 56.26 & 23.31  \\
        & GPT-5 & 16.89 & 54.52 & 54.93 & \textbf{69.10} & \textbf{67.33} & \textbf{68.20} & \textbf{37.46}  \\
        & Claude-4 & 16.60 & 54.02 & 41.37 & 58.53 & 58.44 & 58.48 & 24.19  \\
        & Gemini-2.5 & 18.24 & \textbf{55.67} & 54.87 & \underline{62.65} & \underline{59.03} & \underline{60.79} & 33.35  \\
        \midrule
        \multirow{3}{*}{DCM}
        & DocOwl2(V) & 14.32 & 46.42 & 50.19 & 11.75 & 6.66 & 8.50 & 4.27  \\
        & Docopilot(V) & 16.92 & 53.82 & 39.31 & 18.31 & 17.12 & 17.69 & 6.96  \\
        & Monkey(V) & \textbf{20.16} & 55.19 & 54.61 & 17.08 & 11.27 & 13.58 & 7.41  \\
        \midrule
        \multirow{5}{*}{VLM}
        & GLM-4.6V & \underline{19.38} & 54.76 & \textbf{64.55} & 47.32 & 43.43 & 45.29 & 29.23  \\
        & Qwen3(V) & 14.70 & 53.72 & 22.64 & 20.17 & 20.14 & 20.16 & 4.56  \\
        & GPT-5(V) & 17.32 & 54.85 & 61.47 & 58.90 & 55.34 & 57.07 & \underline{35.08}  \\
        & Claude-4(V) & 13.33 & 50.63 & 31.63 & 54.16 & 53.32 & 53.74 & 16.99  \\
        & Gemini-2.5(V) & 17.27 & 54.85 & 51.71 & 48.39 & 45.59 & 46.95 & 24.28  \\
        \midrule
        \multirow{5}{*}{RAG}
        & HippoRAG2 & 18.71 & 54.16 & 45.77 & 33.13 & 27.88 & 30.28 & 13.86 \\
        & MemoRAG & 13.55 & 52.70 & 51.31 & 24.19 & 19.10 & 21.35 & 10.96 \\
        & Raptor & 18.35 & 54.00 & 36.47 & 25.28 & 20.82 & 22.84 & 8.33 \\
        & VdocRAG(V) & 17.77 & 52.22 & \underline{61.54} & 21.17 & 13.88 & 16.77 & 10.32 \\
        & VisRAG(V) & 16.80 & 54.93 & 39.90 & 26.24 & 23.63 & 24.87 & 9.92 \\
        \bottomrule[1.2pt]
    \end{tabular}
    \caption{Evaluation results of free-form QA on the test set. R-L=ROUGE-L; BERTS.=BERTScore; Concise.=Conciseness.; Correct.=Correctness; Complete. = Completeness; Info. = Informativeness. The best results are highlighted in \textbf{bold}, and the second-best results are \underline{underlined}. 
    }\label{table:main}
\end{table*}

\subsection{Evaluation Protocol} \label{sec:metric}

We establish a unified evaluation framework for assessing academic‑paper understanding.
For binary classification with clear ground-truth labels, 
we use accuracy as the primary metric. 
For open-ended QA, traditional automatic metrics (e.g., BLEU, BERTScore) often fail to capture answer quality since many semantically equivalent responses exist. Following recent work on LLM-as-a-Judge \citep{d2025yescieval,desmond2025evalassist}, we adopt an LLM-based scoring scheme that evaluates each answer along three dimensions: conciseness (brevity without irrelevant content), correctness (accuracy and fidelity, akin to precision), and completeness (coverage of essential content, akin to recall). Each is rated on a 0-5 scale. 
We also compute two derived metrics: an F1-like score (the harmonic mean of correctness and completeness) and informativeness, which aggregates all three dimensions to discourage verbose and repeated outputs (see Appendix \ref{sec:motivation_info} for further motivation).

\begin{equation}\nonumber
\text{F1-like} = 
\frac{(1+\beta^{2}) \times (\text{Correctness} \times \text{Completeness})}
{\beta^{2} \times \text{Correctness} + \text{Completeness}}
\end{equation}

\begin{equation}\nonumber
\text{Informativeness} = \text{F1-like} \times \frac{\text{Conciseness}}{5}
\end{equation}

\noindent where $\beta$ controls the weight between correctness and completeness ($\beta=1$ by default).
This captures the F1-like balance of correctness and completeness, with conciseness penalizing verbosity.

%% file: 4-Experiments.tex
\section{Experiments}

\subsection{Experimental Setup} \label{set:exp_set}

We assess 28 models on the \benchmark\ test set in both text‑only and image‑based settings.
The models span four categories:
\textbf{LLMs}: DeepSeek-V3.2 \citep{liu2025deepseek}, GLM‑4.7 \citep{5team2025glm45agenticreasoningcoding}, Qwen3 (qwen3-235b-a22b) \citep{bai2023qwentechnicalreport}, GPT‑5 (gpt-5-2025-08-07) \citep{leon2025gpt}, Claude‑4 (claude-sonnet-4-20250514) \citep{claude4}, Gemini‑2.5 (gemini-2.5-pro) \citep{comanici2025gemini};
\textbf{Document-Centric Models (DCM)}: DocOwl2(V) \citep{hu2024mplugdocowl2highresolutioncompressingocrfree}, Docopilot(V) \citep{duan2025docopilot}, Monkey(V) \citep{li2024monkeyimageresolutiontext};
\textbf{VLMs}: GLM‑4.6V \citep{vteam2025glm45vglm41vthinkingversatilemultimodal}, Qwen3(V), GPT‑5(V), Claude‑4(V), Gemini‑2.5‑Pro(V);
\textbf{RAG Models}: HippoRAG2 \citep{gutierrez2025rag}, MemoRAG \citep{qian2025memorag}, Raptor \citep{sarthi2024raptor}, VdocRAG(V) \citep{tanaka2025vdocrag}, VisRAG(V) \citep{yu2024visrag}.
Here, “(V)” denotes image‑based input. 
More model results are offered in Appendix \ref{sec:additional_results}. 

As detailed in Section \ref{sec:metric}, open-ended QA is evaluated with ROUGE-L, BERTScore, Conciseness, Correctness, Completeness, F1-like, and Informativeness, while Claim Verification is measured by accuracy. All results are reported on a standardized 0–100 scale. Additional experimental details are provided in Appendix \ref{set:detal_exp_set}.

\subsection{Main Results}

\begin{figure*}[h]
\centering
\includegraphics[width=0.95\linewidth]{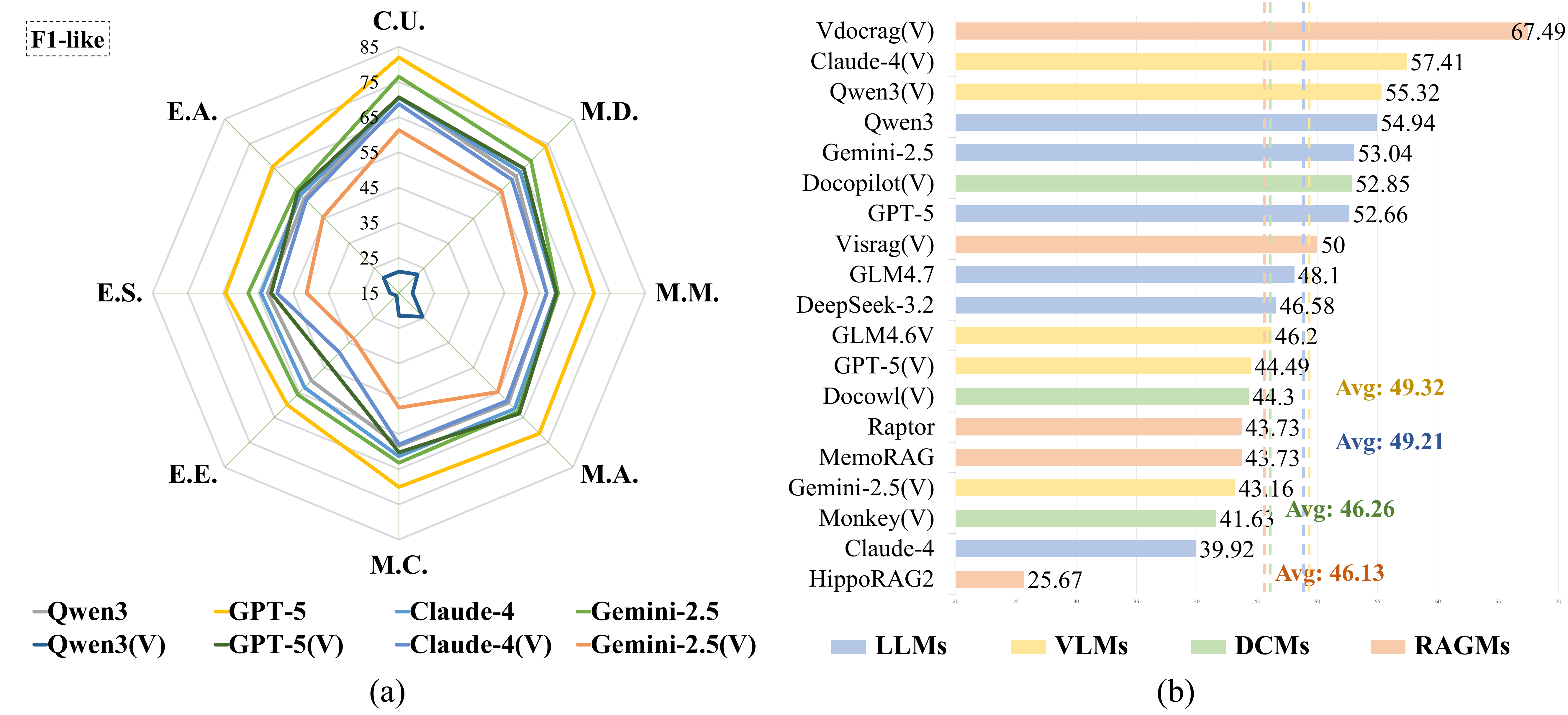}
\caption{Comparison of LLMs and VLMs on open‑ended question answering (F1‑like score; left), and the performance of all models on claim verification tasks (ACC; right).}
\label{fig:task-analysis}
\end{figure*}

Table \ref{table:main} reports all model results on the \benchmark test set, highlighting the following findings:

\textbf{Traditional surface-matching metrics are insufficient for evaluating paper comprehension, as they fail to capture true semantic understanding.} For example, ROUGE-L and BERTScore cannot reliably distinguish large-from small-scale models (LLMs/VLMs vs. DCM/RAG $<$ 10B). 
Monkey(V) attains the best ROUGE-L (20.16\%) and strong BERTScore (55.19\%), yet its correctness and completeness fall to 17.08\% and 11.27\%. 

\textbf{Empirically, LLMs comprehend research papers better through text-only inputs than through images.} Despite multimodal capabilities, 
their high compression ratio forces multimodal models to lose more information.
For example, Qwen3’s F1-like score falls from 56.26\% (text-only) to 20.16\% (image), along with a sharp decline in conciseness. GPT-5(V) appears more concise than its text-only version (61.47\% vs. 54.93\%), but mainly because its responses become shorter and less informative due to reduced correctness and completeness. Overall, the steepest declines occur in correctness and completeness, revealing that current multimodal models still struggle to exploit scholarly visual and textual information.

\textbf{Academic paper comprehension is especially difficult for small models ($\sim$8B).} With limited capacity, document-centric models struggle to integrate information across an entire paper, yielding low F1-like scores (8–18\%) and sometimes incoherent outputs—showing that general-domain fine-tuning is insufficient. 
Poor RAG performance often comes from failing to retrieve the right context, or from smaller models struggling to reason over the retrieved context (See details in Sec. \ref{sec:bottlenecks_of_RAG}).

\textbf{Current models struggle to balance correctness, completeness, and conciseness in paper-based QA.} 
Using Informativeness—a composite metric of these dimensions—we find that even flagship models perform modestly (GPT-5: 37.46\%).
These results point to substantial room for improvement in research paper understanding.

\subsection{Performance across Task Categories}

\paragraph{Performance of Detailed Question Types.}
We evaluate model performance across taxonomy-defined question types. As shown in Figure \ref{fig:task-analysis} (a), models perform better on simpler tasks (e.g., concept understanding, method discrimination) than on deeper reasoning tasks, with the gap widening for image-based inputs. Although figures and tables encode rich information, current models struggle to integrate them with long contexts for coherent reasoning. Overall, most models fail to perform contextual, multimodal reasoning for experimental-specific questions—particularly experimental analysis—highlighting the need for stronger multimodal paper-understanding capabilities.

\paragraph{Results of Claim Verification.} Figure \ref{fig:task-analysis} (b) reports baseline accuracy on claim verification. Overall, the baselines show limited accuracy in claim verification. 
Some multimodal models do relatively well, likely because they better capture a paper’s overall meaning and key claims.
However, large language models may struggle to identify crucial evidence from long contexts. Notably, certain models (e.g., Claude-4, HippoRAG2) show poor instruction-following, often failing to output ``true'' or ``false'' strictly, which further reduces their fact‑checking accuracy.

\subsection{LLM Judgments vs. Human Assessments} \label{sec:llmvshuman}

To align LLM judgments with human assessments, we sample 300 open-ended QA instances from the test set and generate predictions from all models in Section \ref{set:exp_set}. Although prompts limit answers to 3,000 characters (Appendix \ref{sec:lenght_analysis}), actual output lengths vary widely. To control for length effects, for each instance, we select  the three model outputs with the most similar lengths and form pairwise comparisons for annotators, who judge which answer is more correct and complete. During annotation, both presentation order and left–right placement are randomized, and model identities are masked to reduce bias.

\paragraph{Consistency Evaluation Metrics.} We measure consistency between model and human judgments using two metrics: \textbf{BT-based correlation (P-BT, S-BT)}, which fits a Bradley–Terry model~\citep{turner2012bradley} to convert pairwise outcomes into scores and correlates them with human preferences (Pearson/Spearman); and \textbf{pairwise AUC (PW-AUC)}, which directly compares model-predicted pairwise preferences with human labels.

\begin{table}[h]
    \small
    \newcolumntype{C}{>{\centering\arraybackslash}p{2.2em}}
    \centering
    \renewcommand\arraystretch{1.0}
    \begin{tabular}{c|@{~ }*{1}{CCCC}}
            \toprule[1.2pt]
        Setting & P-BT & S-BT & PW-AUC & Avg. \\
        \midrule
        AUG+SEP & 0.8955 & 0.9137 & 0.7125 & \textbf{0.8406} \\
        AUG+JOI & 0.9003 & 0.9091 & 0.6966 & \underline{0.8353} \\
        RAW+SEP & 0.8773 & 0.9137 & 0.7054 & 0.8321 \\
        RAW+JOI & 0.8759 & 0.9137 & 0.7100 & 0.8332 \\
        \bottomrule[1.2pt]
    \end{tabular}
    \captionof{table}{Agreement between human and model judgment w.r.t. prompt configurations. (AUG: enhancement with title and abstract; SEP/JOI: separate/joint evaluation)}
    \label{table:alignment_setting}
\end{table}

\paragraph{Prompt Configuration for LLM Judgments.} We study prompt design via an ablation on two factors: (1) whether the title and abstract are included, and (2) whether to present evaluation metrics separately or jointly. Using GLM-4-Plus as the judge, we test all four combinations. As shown in Table \ref{table:alignment_setting}, including the title and abstract helps the judge understand the questions' context, while assessing each dimension independently mitigates error propagation that can arise when an anomalous score affects multiple dimensions in joint evaluation.

\begin{table}[h]
    \small
    \newcolumntype{C}{>{\centering\arraybackslash}p{2.2em}}
    \centering
    \renewcommand\arraystretch{1.0}
    \begin{tabular}{c|@{~ }*{1}{CCCC}}
            \toprule[1.2pt]
        Model & P-BT & S-BT & PW-AUC & Avg. \\
        \midrule
        $\text{GPT-5}_{int}$ & 0.9059 & 0.9227 & 0.7027 & \underline{0.8437} \\
        $\text{Claude-4.5}_{int}$ & 0.8928 & 0.9136 & 0.7038 & 0.8367 \\
        $\text{Gemini-3}_{int}$ & 0.9091 & 0.9227 & 0.7090 & \textbf{0.8469} \\
        \midrule
        $\text{GPT-5}_{dec}$ & 0.9213 & 0.9137 & 0.7255 & \underline{0.8535} \\
        $\text{Claude-4.5}_{dec}$ & 0.8873 & 0.9091 & 0.7197 & 0.8387 \\
        $\text{Gemini-3}_{dec}$ & 0.9170 & 0.9182 & 0.7330 & \textbf{0.8561} \\
        \bottomrule[1.2pt]
    \end{tabular}
    \captionof{table}{Agreement between different LLM judges and human assessments. Subscript \textit{int} denotes integer scoring, and subscript \textit{dec} denotes decimal scoring.}
    \label{table:alignment_multi_judge}
\end{table}

\paragraph{Which LLMs to Judge?} We first used GLM-4-Plus to score 300 sampled QA instances under the configurations above. From these results, we identified the top three models (GPT-5, Claude-4.5, Gemini-3) and measured their alignment with human assessments (Table \ref{table:alignment_multi_judge}). The two models with the highest alignment were then jointly chosen as evaluation judges to reduce single-judge bias. This procedure is interpretable, extensible, and adaptable to other tasks as resources allow.

\paragraph{Decimal vs. Integer Scoring.} We study two LLM-as-a-Judge scoring schemes: integer scoring and decimal scoring, and test both on the three candidate judge models. As shown in Table \ref{table:alignment_multi_judge}, decimal scoring aligns better with human assessments, whereas integer scoring leads to more clustered scores and reduced sensitivity to subtle but meaningful quality differences. We therefore use decimal scores in our main evaluation.

\subsection{Case Study}\label{sec:appendix_final_cases}

We conduct case studies along four key dimensions: (1) common failure modes of current models (Section \ref{sec:common_failure}), (2) category‑specific analyses (Section \ref{sec:analysis_of_category}), (3) textual versus visual input (Appendix \ref{sec:textual_visual_input}), and (4) bottlenecks of RAG methods (Appendix \ref{sec:bottlenecks_of_RAG}).

\begin{figure*}[t]
    \centering
    \includegraphics[width=0.95\linewidth]{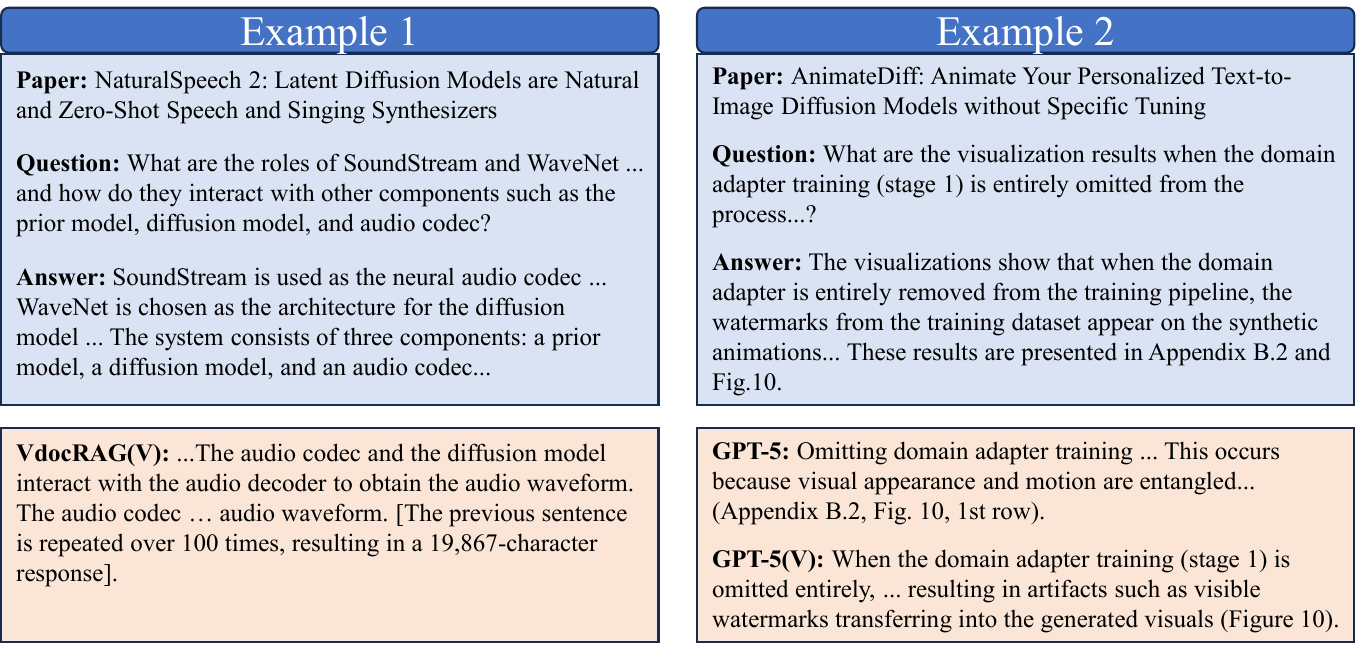}


    \includegraphics[width=0.95\linewidth]{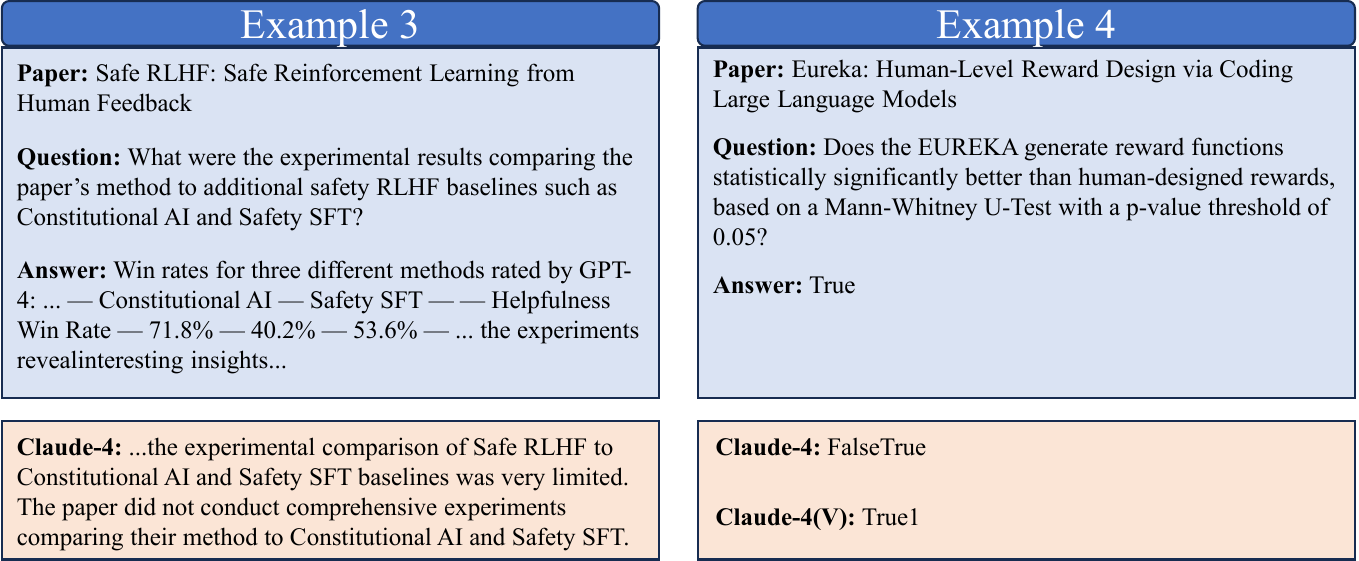}

    \caption{Representative case studies from the RPC‑Bench test set.}
    \label{fig:case-study}
\end{figure*}

\subsubsection{Common Failure Modes} \label{sec:common_failure}
Figure \ref{fig:case-study} illustrates common failure modes of current models in research paper comprehension.

\textbf{Example 1 (Degenerative Output Patterns):} 
The model’s repetitive and uninformative output shows why evaluating conciseness is necessary.

\textbf{Example 2 (Necessity of Multimodal Grounding):} 
Text-only models draw conclusions from text, while multimodal models support claims with visual evidence, showing reasoning capabilities beyond just text analysis.

\textbf{Example 3 (Hallucination):} The model incorrectly denies information that is explicitly present in the source document, highlighting the need for precise evidence extraction and factual verification.

\textbf{Example 4 (Precise Output Failures):} Despite explicit True/False output constraints, both models violate the required format.

\subsubsection{Detailed Analysis of Each Category} \label{sec:analysis_of_category}

We perform a detailed analysis for each of the eight open-ended QA categories.
Detailed examples are provided in Appendix \ref{sec:more_case_category}. 

Across different question types, 
we observe clear and consistent differences among model families.
\begin{itemize}[leftmargin=*]
    \item Advanced LLMs consistently demonstrate strong performance in integrating contextual information, explaining methodological mechanics, and reasoning about experimental design choices, enabling both qualitative and quantitative analysis of results.
    \item VLMs further enhance performance in tasks involving figures or visual evidence but may overlook fine-grained textual details.
    \item RAG systems contribute by ensuring factual accuracy and faithful information extraction, particularly for questions requiring precise retrieval from dispersed sources, yet they generally lack deeper summarization, comparison, and motivation-level reasoning abilities. 
    \item DCMs often give generic, repetitive, or incomplete responses, especially when tasks demand deeper interpretation, comparison, or inference of methodological motivations.

\end{itemize}

%% file: 5-Conclusion.tex
\section{Conclusion}

To comprehensively evaluate models’ ability to understand research papers, we introduce \benchmark, 
a large-scale benchmark with 4,150 research papers and 61.3K QA pairs across 9 categories. 
We develop an LLM–human collaborative annotation framework to ensure scalability and quality. 
The scoring protocol focuses on correctness, completeness, and conciseness, with high-level of agreement between human and model judgments.
Experiments on 28 state-of-the-art models highlight persistent challenges, including limited use of multimodal information, insufficient conciseness, and weak reasoning over visual content. 
\benchmark aims to support testing how well foundation models understand and reason about research papers.

%% file: 6-Appendix.tex
\section{Supplementary Experiments}

\subsection{Detailed Experimental Settings} \label{set:detal_exp_set}

We evaluate models across the two configurations: pure-text and image-based. For text, each PDF is converted to Markdown via MinerU\footnote{https://github.com/opendatalab/MinerU}
, with content truncated if it exceeds the model’s context window. For images, PDFs are rendered with PyMuPDF at 200 DPI, and the first 15 pages are used to balance coverage and context limits. For models without multi-image support (e.g., Monkey), these pages are concatenated into a single composite image for compatibility.
It is important to note that limitations such as text truncation and image constraints during inference stem from the current capabilities of LLMs/VLMs rather than from the design of our benchmark. The dataset itself includes the complete text and image content.

To maximize performance while leveraging each model’s strengths, we impose minimal inference constraints: (1) answers must rely only on the given paper; (2) open-ended responses must be professional, concise, and under 3,000 characters; (3) claim verification outputs must be strictly True or False. 
The complete prompts for the two task types are listed in Appendix \ref{sec:prompt_reasoning}, and the full evaluation prompts are included in Appendix \ref{sec:prompt_evaluation}.

\subsection{Main Results with Additional Models} \label{sec:additional_results}

We further evaluate 9 models on the \benchmark\ test set under two input configuration (pure-text and image-based) to provide a more comprehensive comparison: 
\textbf{LLMs}: DeepSeek-V3.1, GLM‑4.5, GPT‑5.2 (gpt-5.2-2025-12-11), Claude‑4.5 (claude-sonnet-4-5-20250929), Gemini‑3 (gemini-3-pro-preview-11-2025);
\textbf{VLMs}: GLM‑4.5V, GPT‑5.2(V), Claude‑4.5(V), Gemini‑3(V).
The results are summarized in Table \ref{table:additional_main}. 

\begin{table*}[t]
    \newcolumntype{C}{>{\centering\arraybackslash}p{2.2em}}
    \centering
    \renewcommand\arraystretch{1.0}
    \begin{tabular}{c|c|@{~ }*{1}{CC|}*{1}{CCCCC}}
            \toprule[1.2pt]
        \multirow{2}{*}{\shortstack{Model\\Type}}    
            &\multirow{2}{*}{Model } 
            &\multicolumn{2}{c|}{Traditional}
            &\multicolumn{5}{c}{LLM-as-judge}
        \\
        \cmidrule{3-4} \cmidrule{5-9}
        & & {R-L} & {B-S} & {Conc.} & {Corr.} & {Compl.} & \mbox{F1-like} & {Info.} \\
        \midrule
        \multirow{6}{*}{LLM}
        & DeepSeek-V3.1 & \underline{19.12} & \textbf{55.98} & 54.76 & 57.85 & 54.85 & 56.31 & 30.84  \\
        & GLM-4.5 & 16.03 & 53.18 & 43.41 & 58.95 & 59.54 & 59.24 & 25.72  \\
        & GPT-5.2 & 16.90 & 54.00 & 53.81 & \textbf{66.84} & \underline{64.03} & \textbf{65.40} & \textbf{35.19}  \\
        & Claude-4.5 & 12.75 & 50.62 & 31.02 & \underline{64.31} & \textbf{64.97} & \underline{64.64} & 20.05  \\
        & Gemini-3 & 17.74 & 55.14 & 52.81 & 62.69 & 60.28 & 61.46 & \underline{32.46}  \\
        \midrule
        \multirow{5}{*}{VLM}
        & GLM-4.5V & \underline{19.66} & \underline{55.48} & \textbf{59.44} & 48.79 & 43.62 & 46.06 & 27.38  \\
        & GPT-5.2(V) & 16.80 & 53.97 & \underline{56.43} & 56.75 & 52.82 & 54.72 & 30.88  \\
        & Claude-4.5(V) & 13.47 & 51.19 & 31.95 & 55.35 & 54.45 & 54.89 & 17.54  \\
        & Gemini-3(V) & 16.60 & 54.38 & 50.22 & 56.06 & 52.69 & 54.32 & 27.28  \\
        \bottomrule[1.2pt]
    \end{tabular}
    \caption{Evaluation results of free-form QA on the test set. RG-L=ROUGE-L; B-S=BERTScore-F1; Compl. = Completeness; Info. = Informativeness. The best results are highlighted in \textbf{bold}, and the second-best results are indicated with \underline{underlining}. 
    }\label{table:additional_main}
\end{table*}

\subsection{Response Length Analysis} \label{sec:lenght_analysis}

To better understand output length characteristics, we analyzed the distribution of response lengths for all models, as summarized in Table \ref{table:length}. 
However, most models tended to produce responses approaching the upper bound, which we attribute to their limited ability to comprehend and reason over the research paper content, leading to verbose rather than concise answers. 
Notably, some models (such as DocOwl2, VdocRAG, and VisRAG) exhibited abnormally high maximum output lengths. Manual inspection revealed that these overly long outputs often consisted of repetitive, non‑informative text generated when the model failed to answer the question effectively. Conversely, certain models registered a minimum response length of zero, indicating empty answers either due to refusal triggered by safety policies (e.g., GLM‑4.5V) or an inability to provide a response.
Overall, the most models struggle to effectively achieve content comprehension, information compression, and logical reasoning within the task constraints, revealing a fundamental gap between the demands of accurate, concise, and contextually grounded scholarly reasoning and the current capabilities of state‑of‑the‑art systems.

\begin{table*}[h]
    \newcolumntype{C}{>{\centering\arraybackslash}p{4em}}
    \centering
    \renewcommand\arraystretch{1.0}
    \begin{tabular}{c|c|@{~ }*{1}{CCC}}
            \toprule[1.2pt]
        \multirow{2}{*}{Model Type}     
            &\multirow{2}{*}{Model} 
            &\multicolumn{3}{c}{Length (char)}
        \\
        \cmidrule{3-5}
        & & {Avg} & {Max.} & {Min}\\
        \midrule
        \multirow{6}{*}{LLM}
        & DeepSeek-V3.1 & 1494.19 & 4350 & 51  \\
        & DeepSeek-V3.2 & 1508.72 & 3814 & 1  \\
        & GLM-4.5 & 2127.67 & 10285 & 478  \\
        & GLM-4.7 & 1209.72 & 4175 & 0  \\
        & Qwen3 & 1971.16 & 5527 & 289  \\
        & GPT-5 & 1717.14 & 4065 & 37  \\
        & GPT-5.2 & 1809.17 & 4202 & 183  \\
        & Claude-4 & 2109.62 & 5198 & 85  \\
        & Claude-4.5 & 3925.48 & 8714 & 13  \\
        & Gemini-2.5 & 1741.91 & 3985 & 330  \\
        & Gemini-3 & 1928.55 & 4153 & 0  \\
        \midrule
        \multirow{3}{*}{DCM}
        & DocOwl2 & 525.52 & 23251 & 0  \\
        & Docopilot & 1108.01 & 4031 & 59  \\
        & Monkey & 425.10 & 7351 & 10  \\
        \midrule
        \multirow{5}{*}{VLM}
        & GLM-4.5V & 1090.89 & 4079 & 0  \\
        & GLM-4.6V & 893.45 & 3154 & 0  \\
        & Qwen3(V) & 2019.61 & 7627 & 249  \\
        & GPT-5(V) & 1332.63 & 4698 & 24  \\
        & GPT-5.2(V) & 1536.19 & 4268 & 140  \\
        & Claude-4(V) & 3530.55 & 9853 & 821  \\
        & Claude-4.5(V) & 3363.52 & 8205 & 611  \\
        & Gemini-2.5(V) & 1667.97 & 3256 & 274  \\
        & Gemini-3(V) & 1962.49 & 3880 & 0  \\
        \midrule
        \multirow{5}{*}{RAG}
        & HippoRAG2 & 1109.48 & 3011 & 39 \\
        & MemoRAG & 594.26 & 1957 & 1 \\
        & Raptor & 838.01 & 1956 & 0 \\
        & VdocRAG & 2391.54 & 30179 & 1 \\
        & VisRAG & 1232.17 & 13046 & 2 \\
        \midrule
        \multirow{1}{*}{-}
        & Ground True & 829.63 & 5194 & 5 \\
        \bottomrule[1.2pt]
    \end{tabular}
    \caption{Response Length Analysis.}
    \label{table:length}
\end{table*}

\subsection{Finetune LLM Analysis}

We fine‑tuned Qwen and LLaMA on the PRC training set, with results summarized in Table \ref{table:finetune}. Both models achieved consistent improvements in the overall Info. metric, increasing by 11.38\% and 10.64\%, respectively. Notably, conciseness improved significantly, whereas the F1‑like score remained relatively stable. This suggests that, compared to correctness and completeness, models more readily learn to produce concise responses. In contrast, achieving high correctness and completeness imposes greater demands on the models’ fundamental comprehension and reasoning capabilities.

\begin{table*}[h]
    \newcolumntype{C}{>{\centering\arraybackslash}p{4em}}
    \centering
    \renewcommand\arraystretch{1.0}
    \begin{tabular}{c|@{~ }*{1}{CCCCC}}
            \toprule[1.2pt]
        \cmidrule{2-6}
        Model Type & {Conc.} & {Corr.} & {Compl.} & {F1-like} & {Info.}\\
        \midrule
        Llama-3.1-8B-Instruct & 41.56 & 34.75 & 30.92 & 32.72 & 13.60  \\
        Llama-3.1-8B-Instruct-FT & 77.07 & 36.20 & 29.34 & 32.41 & 24.98  \\
        Qwen3-8B & 48.20 & 38.53 & 32.52 & 35.27 & 17.00  \\
        Qwen3-8B-FT & 78.58 & 39.30 & 31.82 & 35.17 & 27.64  \\
        \bottomrule[1.2pt]
    \end{tabular}
    \caption{Performance Comparison in Fine‑Tuning Experiments.}
    \label{table:finetune}
\end{table*}

\subsection{Evaluation of Model Conciseness across Taxonomy‑Defined Question Types}

\begin{figure*}[t]
\centering
\includegraphics[width=0.75\linewidth]{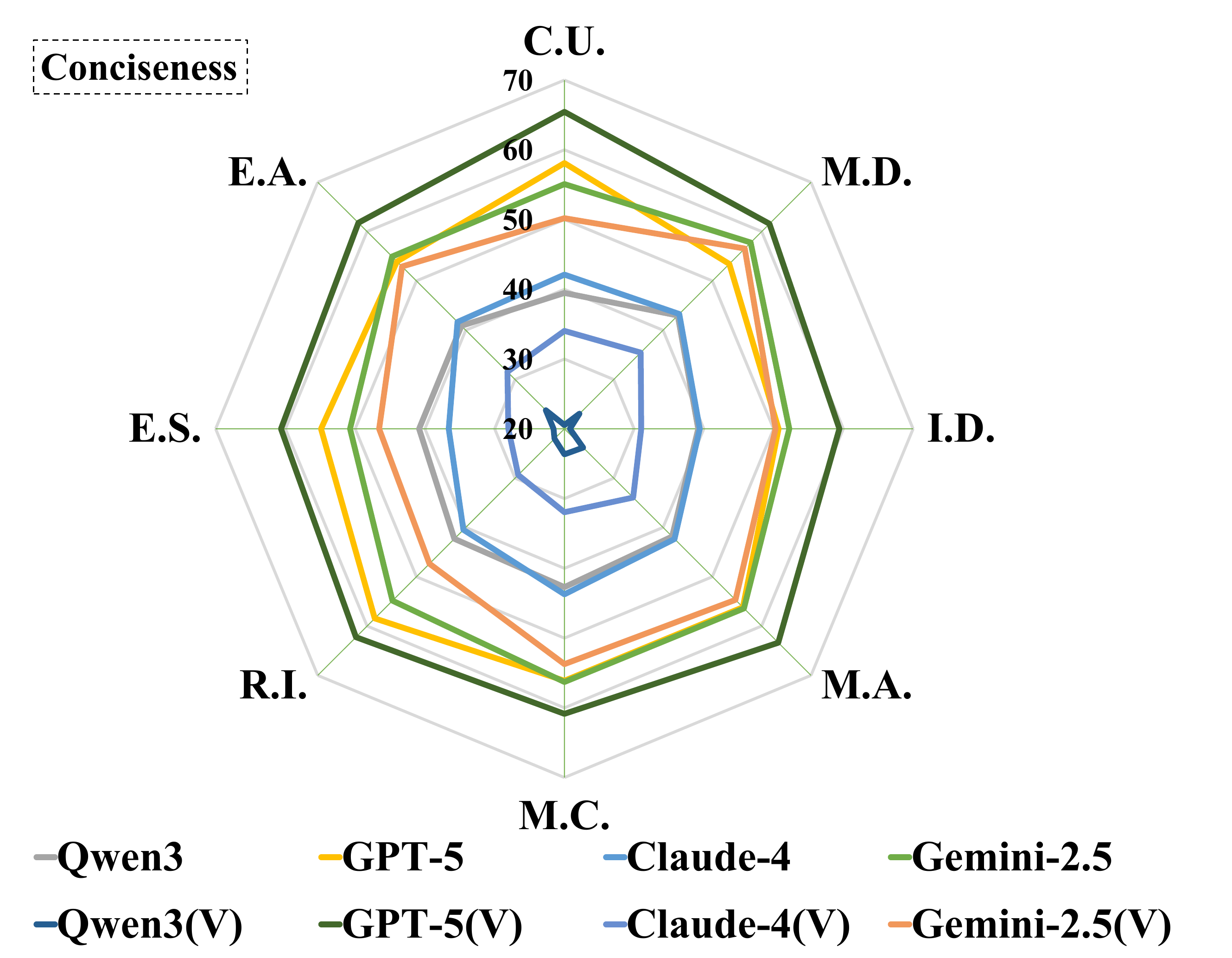}
\caption{Model Conciseness across Taxonomy‑Defined Question Types}
\label{fig:conciseness}
\end{figure*}

Figure \ref{fig:conciseness} illustrates the conciseness of different models across various question types. 
Most models show minimal variation in conciseness scores across question categories, forming an almost concentric pattern in the radar chart, with no score exceeding 65.5\%. This underscores the difficulty models face in generating responses that are both relevant and precise. Across all question categories, text-based inputs generally yield more concise outputs than image-based inputs. We attribute this to the models’ weaker capability in interpreting visual inputs, where the relative loss of explicit textual detail may result in responses that convey less relevant information while being unnecessarily verbose. This pattern is evident in the performance gap between text- and image-based inputs for Qwen3, Gemini-2.5, and Claude-4. An exception is GPT‑5(V), which achieves the highest answer conciseness, even surpassing its text-based variant. This result suggests that GPT‑5(V) can more effectively leverage visual information, consistent with its strong F1‑Like score (see Figure \ref{fig:task-analysis} (left)).

\subsection{More Case Study} \label{sec:more_case}

\subsubsection{Detailed Analysis of Each Category} \label{sec:more_case_category}

Representative examples for each category are presented below.

\vspace{0.2cm}

\begin{exampleblock}[Concept Understanding]

\textbf{Question:} What is "cross-LoRA attention" as defined in the paper?
\\\\
\textbf{Answer:} Cross-LoRA attention is an attention mechanism that allows for combining concept and style LoRAs. It splits queries, keys, and values, such that queries from the concept branch attend to keys and values from both the concept and style branches, effectively merging their representations.
\\\\
\textbf{GPT:} Cross-LoRA attention is an attention mechanism designed to combine the concept and style LoRAs... It involves splitting the queries, keys, and values (Q, K, V) into concept and style components. The concept queries attend to the concatenated keys and values from both the concept and style branches...
\\\\
\textbf{GPT(V):} Cross-LoRA attention is the mechanism the authors propose to merge the learned representations of a concept LoRA and a style LoRA during inference... The key idea is to take the query from the concept branch and let it attend to the keys and values from both the concept and style branches.
\\\\
\textbf{DocOwl2:} Cross-LoRA attention is a novel method that enables the model to focus on different aspects of the input image, allowing for more detailed and accurate image generation.
\\\\
\textbf{VisRAG:} Cross-LoRA attention is defined as a mechanism for merging LoRA weights by splitting the query (q) into a concept part (qc) and a style part (qs), and then having the concept query attend to keys and values from both branches.
\end{exampleblock}

\vspace{0.2cm}

\begin{exampleblock}[Method Disambiguation]
\textbf{Question:} How does the positional embedding used in Rectified Flow Transformers (RFTs) differ from the one used in Diffusion Transformers (DiTs)?
\\\\
\textbf{Answer:} RFTs condition on a continuous time variable `t` from `[0, 1]`, while DiTs condition on a discrete noise level `$\Sigma$`. This allows RFTs to leverage the properties of rectified flow for training.
\\\\
\textbf{Claude-4:} The key difference is that RFTs condition on the continuous time variable 't' which ranges from 0 to 1... In contrast, DiTs condition on discrete noise levels '$\Sigma$'.
\\\\
\textbf{Qwen3(V):} Both RFTs and DiTs utilize positional embeddings... However, the paper does not explicitly detail the structural differences... it focuses on the Rectified Flow method itself rather than a component-wise comparison of embeddings.
\\\\
\textbf{DocOwl2:} The positional embedding in RFTs is based on the transformer architecture, while the positional embedding in DiTs is based on the diffusion model architecture.
\\\\
\textbf{HippoRAG2:} Rectified Flow Transformers (RFT) condition on time `t`, whereas DiTs condition on the noise level `$\Sigma$`.
\end{exampleblock}

\vspace{0.2cm}

\begin{exampleblock}[Method Mechanics]
\textbf{Question:} How is normalization applied in the description of Classification Accuracy?
\\\\
\textbf{Answer:} Normalization is performed over the multi-choice options given, rather than over all sequences in the vocabulary.
\\\\
\textbf{Claude-4:} The key aspect of normalization is that it is "performed over the options given, rather than over all sequences." This means that instead of normalizing probabilities across the entire vocabulary... the normalization is restricted to only the specific answer choices provided...
\\\\
\textbf{Claude-4(V):} ...the model uses the probability of the full token sequences corresponding to each class. These probabilities are renormalized to sum to 1 over the provided options for the task... It is done **over the given options only**, rather than over all possible sequences in general...
\\\\
\textbf{Monkey:} Normalization is applied... by dividing the number of correctly classified instances by the total number of instances in the test set.
\\\\
\textbf{MemoRAG:} Normalization is done over the options given, rather than over all sequences.
\end{exampleblock}

\vspace{0.2cm}

\begin{exampleblock}[Motivation Analysis]
\textbf{Question:} Why was RAFT not tested on tasks such as text-to-SQL or reasoning-based QA, given its potential effectiveness in those domains?
\\\\
\textbf{Answer:} RAFT is specifically designed to enhance the base LLM's performance in domain-specific RAG settings. While it might improve performance in coding or reasoning domains, studying RAFT's performance on those tasks falls outside the scope of this paper, as it focuses on creating domain expert RAG models rather than general post-training strategies.
\\\\
\textbf{Claude-4:} ...RAFT was not tested on text-to-SQL or reasoning-based QA tasks because the authors made specific dataset choices aligned with their research objectives and scope limitations... The paper explicitly states their dataset selection criteria...
\\\\
\textbf{Claude-4(V):} ...due to the focus and scope of this paper, which targets domain-specific open-book question answering (QA) settings... RAFT’s method depends on training with **golden and distractor documents**... [This] is fundamentally different from generating executable code (SQL queries)...
\\\\
\textbf{Monkey:} The paper does not provide a direct reason...
\\\\
\textbf{MemoRAG:} RAFT was not tested on... text-to-SQL or reasoning-based QA because the paper's focus is specifically on adapting pre-trained LLMs to domain-specific retrieval-Augmented Generation (RAG) in specialized domains.
\end{exampleblock}

\vspace{0.2cm}

\begin{exampleblock}[Method Comparison]
\textbf{Question:} How does Self-RAG differ from other RAG methods, such as those that use iterative retrieval or reranking?
\\\\
\textbf{Answer:} Self-RAG differs by adaptively retrieving passages on-the-fly and using self-reflection. Unlike iterative retrieval, it retrieves when needed during generation. Unlike reranking, which happens before generation, Self-RAG’s critiques happen post-generation to select the best output. It generates "reflection tokens" to make these decisions.
\\\\
\textbf{Gemini-2.5:} Self-RAG differs by training a single language model to adaptively retrieve passages on an on-demand basis and to **self-reflect on the retrieved passages**... Unlike iterative retrieval which retrieves a fixed number of documents... Self-RAG retrieves passages whenever it deems necessary... Unlike reranking... Self-RAG uses its own critique mechanism after generation...
\\\\
\textbf{Gemini-2.5(V):} Self-RAG differs in that it is an on-demand retrieval system that works during the decoding process... It also generates special tokens (`[Retrieve]`, `[Critique]`) that allow it to assess its own generation, unlike other methods that re-rank at the start.
\\\\
\textbf{Monkey:} Traditional RAG methods retrieve information, then generate a response. Self-RAG adds a "critique" step where the model evaluates its own generation. However, the paper does not compare it to iterative retrieval.
\\\\
\textbf{HippoRAG2:} Self-RAG is different because it trains one arbitrary LM to generate text and reflect on its own output... by generating special tokens... It retrieves passages one at a time and in an adaptive manner, unlike conventional RAG that retrieves a fixed number of passages.
\end{exampleblock}

\vspace{0.2cm}

\begin{exampleblock}[Experimental Exposition]
\textbf{Question:} What are the effects of not including a weight threshold, and are there specific tasks where this negatively impacts the results?
\\\\
\textbf{Answer:} An ablation study showed minimal impact on most tasks. However, three specific tasks-'Date Understanding,' 'Disambiguation,' and 'Hyperbaton'-showed notable performance declines, with an average decrease of 1.2\%.
\\\\
\textbf{GPT-5:} Effect on overall performance: Removing the weight threshold... has minimal impact on most tasks... reduces the average performance across BBH from 34.7 to 33.5... Tasks negatively impacted: Three tasks show notable drops... Date Understanding: 32.9 → 17.7... Disambiguation: 45.2 → 40.6... Hyperbaton: 62.8 → 55.5...
\\\\
\textbf{GPT-5(V):} The paper does not study or report a “weight threshold” mechanism.**... Consequently: - The effects of not using a weight threshold are not evaluated or quantified in the paper. - No tasks are identified where omitting a weight threshold harms performance.
\\\\
\textbf{DocOwl2:} ...including a weight threshold does not significantly affect the method's performance, and there are no specific tasks where this adjustment negatively impacts the results.
\\\\
\textbf{VisRAG:} The removal of the threshold had minimal impact on the majority of tasks... However, three tasks namely Date Understanding, Disambiguation and Hyperbaton exhibited notable effect... average decrease of 1.2\%...
\end{exampleblock}

\vspace{0.2cm}

\begin{exampleblock}[Experimental Setup]
\textbf{Question:} Can the success/fitness function be used to initialize the Eureka reward search process?
\\\\
\textbf{Answer:} Yes, the success/fitness function can be used. Section 4.4 describes an experiment where a human - supplied reward function was used to initialize the process.
\\\\
\textbf{Gemini-2.5:} Yes, the success/fitness function can be used to initialize the Eureka reward search process, although it is not the default procedure... Section 4.4 explains that initializing the search with an existing reward function "requires no modification to EUREKA."
\\\\
\textbf{Gemini-2.5(V):} Based on the provided research paper, the success/fitness function isnotused to initialize the Eureka reward search process... The initialization of the reward search process involves providing the Large Language Model (LLM) with two key pieces of information as context: 1. The environment source code. 2. A natural language task description.
\\\\
\textbf{Monkey:} Yes, the success/fitness function can be used to initialize the Eureka reward search process.
\\\\
\textbf{MemoRAG:} Yes, the success/fitness function can be used to initialize the Eureka reward search process.
\end{exampleblock}

\vspace{0.2cm}

\begin{exampleblock}[Experimental Analysis]
\textbf{Question:} What are the major errors identified in the error analysis conducted in Section 5.1, and how do these errors manifest in the model's predictions?
\\\\
\textbf{Answer:} The major errors identified in the error analysis lie within the click region and scroll direction predictions. The model tends to click the wrong place or scroll in the wrong direction, despite predicting the right action most of the time. This suggests a need for improving the model’s ability to understand screen layouts.
\\\\
\textbf{GPT-5:} Section 5.1 identifiesclick region localizationandscroll direction predictionas the major error sources... These errors manifest as the model selecting the correct type of action but failing at the low - level execution...
\\\\
\textbf{GPT-5(V):} Section 5.1 shows that Auto - UI’s main errors occur at the execution level, not in deciding what to do. Error types:inaccurate click localizationandincorrect scroll direction... Evidence: averaged category accuracies are 67.4\% for click region and 82.0\% for scroll direction...
\\\\
\textbf{DocOwl2:} The model's predictions are not accurate because of overfitting and underfitting... identified as model drift and model error...
\\\\
\textbf{MemoRAG:} Major errors include incorrect location prediction during clicks (8\%) and misdirection during scroll movements (4\%).
\end{exampleblock}

\subsubsection{Textual vs. Visual Input} \label{sec:textual_visual_input}

This section provides a qualitative comparison of model behavior under purely textual and visual inputs. 

Example 9, 10, and 12 shows that LLMs consistently deliver higher factual accuracy, greater answer completeness, and closer alignment with reference answers than VLMs.
In Example 11, VLMs accurately identified methodological boundaries (e.g., excluding tasks outside the intended domain‑specific RAG scope) and maintained reasoning consistency with the reference. This indicates that VLMs can effectively capture high‑level conceptual scope, especially when the task focuses on boundary recognition rather than detailed fact retrieval.

Overall, LLMs excel in tasks requiring precise, evidence‑linked, detail‑rich answers from textual content. VLMs occasionally match or surpass LLMs in identifying conceptual scope from visual inputs. The observed gap largely reflects VLMs’ difficulty in extracting high‑precision information from PDF‑rendered images of dense academic content (e.g., complex tables, multi‑column layouts, long texts), which often results in incomplete or noisy answers compared with clean text inputs.

\vspace{0.2cm}

\begin{exampleblock}[]
\textbf{Question:} Can the success/fitness function be used to initialize the Eureka reward search process?
\\\\
\textbf{Answer:} Yes, the success/fitness function can be used. Section 4.4 describes an experiment where a human-supplied reward function was used to initialize the process.
\\\\
\textbf{GPT-5:} Yes. EUREKA can be initialized from any provided reward function by substituting it for the first iteration’s reward, without modifying the algorithm. The paper demonstrates this “initialization” mechanism with human-written rewards: “we can simply substitute the raw human reward function as the output of the first EUREKA iteration” (Sec. 4.4; App. D.2). In the benchmarks, the success/fitness function F is provided as executable reward code and is used directly as the “Sparse” reward baseline (“These are identical to the fitness functions F…,” Sec. 4.1; App. B). Therefore, F can be used in exactly the same way as an initialization for EUREKA: set the first iteration’s reward to F, evaluate it, and then let EUREKA’s evolutionary search and reward reflection refine it in subsequent iterations (Alg. 1; Sec. 3.2–3.3).
\\\\
\textbf{GPT-5(V):} Yes, the success/fitness function ( F ) can be used to initialize the Eureka reward search process, as outlined in the paper. Eureka explicitly uses ( F ) to evaluate the quality of candidate reward functions generated during its iterative optimization process (Section 3.2). ( F ) measures how effectively a policy derived from a reward function ( R ) solves a task (Definition 2.1). In the initial phase, the fitness function is employed to assess the performance of reward functions sampled zero-shot from the coding LLM, enabling an informed starting point for the evolutionary search process (Section 3.1). While the initial rewards may exhibit suboptimality relative to the fitness metric, this evaluation ensures that the process begins with executable and task-relevant reward functions, which are progressively refined in subsequent iterations using evolutionary search and reward reflection. Thus, ( F ) plays a critical foundational role in initializing and guiding Eureka's reward optimization.
\end{exampleblock}

\vspace{0.2cm}
\begin{exampleblock}[]
\textbf{Question:} What are the details of LoRA fine-tuning, including the rank, learning rate, and model architecture used?
\\\\
\textbf{Answer:} The details of LoRA fine-tuning include incorporating trainable linear low-rank modules into the query and value projections of each self-attention layer, following Hu et al. (2021). The rank of LoRA is set to 32, the learning rate is $1e-4$, and the model architecture is the Gecko PaLM model.
\\\\
\textbf{GPT-5:} LoRA fine-tuning details (as used in SQL-PaLM): - Placement: Trainable low-rank adapters inserted into the query and value projections of each self-attention layer. - Rank: 32.- Learning rate: $1e-4$. - Model architecture: PaLM-2 Gecko (Gecko PaLM model).
\\\\
\textbf{GPT-5(V):} The LoRA (Low-Rank Adaptation) fine-tuning in the SQL-PaLM framework employs the following attributes: 1. Rank: LoRA fine-tuning utilizes a rank of 64, representing the dimensionality of the low-rank decomposition matrices used to adapt large model weights. 2. Learning Rate: A fixed learning rate of 5e-4 is used for the training process with LoRA. 3. Model Architecture: LoRA is applied to the PaLM-2-Large model, a pretrained large-scale Transformer-based architecture. These details align with the authors' exploration of parameter-efficient fine-tuning techniques tailored for scaling Text-to-SQL task performance.
\end{exampleblock}

\vspace{0.2cm}

\begin{exampleblock}[]
\textbf{Question:} Why was RAFT not tested on tasks such as text-to-SQL or reasoning-based QA, given its potential effectiveness in those domains?
\\\\
\textbf{Answer:} RAFT is specifically designed to enhance the base LLM's performance in domain-specific RAG settings. While it might improve performance in coding or reasoning domains, studying RAFT's performance on those tasks falls outside the scope of this paper, as it focuses on creating domain expert RAG models rather than general post-training strategies.
\\\\
\textbf{GPT-5:} Two points from the paper address this: - Reasoning-based QA: RAFT was evaluated on a reasoning-intensive QA benchmark. HotpotQA-an explainable, multi-hop QA dataset-is one of the core evaluations, and RAFT shows large gains there (e.g., 35.28 vs. much lower baselines in Table 1; further analysis in Sections 4.1–4.3). Thus, RAFT was tested on reasoning-based QA within the paper’s scope. - Text-to-SQL: The paper’s stated focus is domain-specific, document-centric RAG: “we only care about testing the LLM on the same set of documents,” and “the LLM will be used to respond to the questions, whose answers can be found within a collection of documents.” All methods, data construction, and evaluations are built around reading retrieved documents, citing verbatim spans, and being robust to distractor documents. The chosen datasets (Wikipedia-based QA, API documentation, PubMed) reflect this scope. Tasks like text-to-SQL are not part of this document-retrieval setting the paper studies, so they were outside the evaluation slate.
\\\\
\textbf{GPT-5(V):} Because the paper’s goal and problem setting are narrowly defined as domain‑specific open‑book QA with retrieval. The method and experiments are built around: - Training and testing with a question plus a set of retrieved documents, where “answers can be found within a collection of documents” in the target domain (Domain‑Specific Open‑Book Exam section). - RAFT’s training format requires a golden document D* that contains the answer and supports chain‑of‑thought with verbatim citations from the relevant document (Section 3). - The evaluation is designed to study reading from documents and robustness to distractors in RAG, using datasets such as PubMed, HotPotQA, and Gorilla APIBench that match this setting (Section 4). Tasks like text‑to‑SQL or pure reasoning‑based QA do not align with this document‑grounded RAG setup and do not provide the golden supporting document and citations RAFT relies on. Hence, they were outside the scope of the study and were not evaluated.
\end{exampleblock}

\vspace{0.2cm}

\begin{exampleblock}[]
\textbf{Question:} How does WizardLM address instructions that are not included in the evolved set?
\\\\
\textbf{Answer:} WizardLM addresses instructions not included in the evolved set through In-Breadth Evolving, which evolves entirely new instructions to enhance topic coverage, skill coverage, and overall diversity. This approach is demonstrated in Section 4.5 and Appendix J, showing improved diversity in the evolved instruction data compared to the original data.
\\\\
\textbf{GPT-5:} WizardLM handles missing or failed evolutions in two ways: - During evolution: If an instruction fails to produce a valid evolved version (per the four failure checks), the evolved sample is discarded and the original instruction is put back into the pool “as is,” to be attempted again in subsequent evolution epochs. - For training data assembly: After all rounds, the final fine-tuning set is the merge of the initial instruction dataset and all successfully evolved instructions. Thus, instructions that never appear in the evolved set are still included via their original form. Additionally, WizardLM uses In-breadth Evolving to create brand-new, rarer instructions from existing ones, expanding topic and skill coverage to mitigate gaps where specific instructions are not represented in the evolved set.
\\\\
\textbf{GPT-5(V):} WizardLM does not discard instructions that are not part of the evolved set. - During evolution, if an instruction fails to evolve (e.g., gives no information gain, is hard to answer, yields trivial output, or copies words from the prompt), it is “placed back as it is,” to be attempted again in later evolution rounds. - When building the final fine-tuning data, the authors merge the initial instruction dataset with all evolved instructions and shuffle them, ensuring unevolved original instructions are still included in training. Thus, instructions not included in the evolved set are retained and reused, both for further evolution attempts and by inclusion in the final fine‑tuning corpus.
\end{exampleblock}

\subsubsection{Bottlenecks of RAG Methods} \label{sec:bottlenecks_of_RAG}

We conducted a further analysis of the RAG-based approach, focusing on representative cases:

Example 13: The model successfully retrieved information related to the question but failed to use it effectively in answer generation. Key links between the paper’s context and the question were overlooked, and model‑generated hallucinations were introduced. 

Example 14: The model failed to retrieve the key information needed to answer the question, such as the main innovations of the method. Without this critical material, the generated answer was incomplete or incorrect.

Example 15: The model broke the question down into sub‑questions, retrieved all necessary content, and produced a coherent, complete answer.

Overall, The main bottleneck of current RAG‑based approach lies in their limited ability to accurately understand complex questions, retrieve targeted knowledge, and integrate information from multiple sources into a unified, meaningful answer. 

\vspace{0.2cm}

\begin{exampleblock}[]

\textbf{Question:} Under what conditions does the statement $\mathcal G_\text{g} \geq 0$ hold, and what could lead to $\mathcal G_\text{g} < 0$?
\\\\
\textbf{Answer:} The statement $\mathcal G_\text{g} \geq 0$ holds under the assumption that training and test data come from the same distribution. If this assumption is violated and the test data is simpler (i.e., from a distribution with lower entropy) than the training data, it can lead to $\mathcal G_\text{g} < 0$.
\\\\
\textbf{Retrieve:} and the gap $\mathcal { G } _ { \mathrm { r } }$ is a non-negative value

Note that $p _ { \mathrm { a u g } } ( { \pmb x } ^ { \prime } ) = \mathbb { E } _ { { \pmb x } \sim \mathcal { D } _ { \mathrm { t r a i n } } } [ p _ { \mathrm { a u g } } ( { \pmb x } ^ { \prime } \mid { \pmb x } ) ]$ only extrapolates from individual data points $x \sim \mathcal { D } _ { \mathrm { t r a i n } }$ and has density outside the support of $p _ { \mathrm { d a t a } } ( \pmb { x } )$ (e

$\mathcal{G}_{\mathrm{a}} = \mathbb{E}_{\mathbf{x} \sim \mathcal{D}_{\mathrm{test}}}[\mathrm{ELBO}^{*}_{\theta}(\mathbf{x})] - \mathbb{E}_{\mathbf{x} \sim \mathcal{D}_{\mathrm{test}}}[\mathrm{ELBO}_{\Theta}(\mathbf{x})]$

the ELBO on $\mathcal { D } _ { \mathrm { t r a i n } }$ is greater than or equal to the ELBO on $\mathcal { D } _ { \mathrm { t e s t } }$

Therefore the ELBO on $\mathcal { D } _ { \mathrm { t e s t } }$ can be higher than the ELBO on $\mathcal { D } _ { \mathrm { t r a i n } }$ if $\mathcal { D } _ { \mathrm { t r a i n } }$ and $\mathcal { D } _ { \mathrm { t e s t } }$ are not drawn from the same distribution

and $\mathcal { D } _ { \mathrm { t e s t } }$ has a lower entropy than $\mathcal { D } _ { \mathrm { t r a i n } }$

 Hence, the continuity we gain by replacing $\mathcal { D } _ { \mathrm { t r a i n } }$ with $p _ { \mathrm { D M } } ( \pmb { x } ^ { \prime } )$ is not for free, we lose a small amount of information about $\mathcal { D } _ { \mathrm { t r a i n } }$ 

\[
\mathcal{G}_{\mathrm{r}} =
\mathbb{E}_{\mathbf{x}^{\mathrm{r}} \sim \mathcal{D}_{\mathrm{test}}}
\; \mathbb{E}_{\mathbf{x}^{\mathrm{a}} \sim p(\mathbf{x}^{\mathrm{a}} \mid \mathbf{x}^{\mathrm{r}})}
\left[ \mathrm{MS\!-\!SSIM} \left( \mathbf{x}^{\mathrm{r}}, \mathbf{x}^{\mathrm{a}} \right) \right]
\]

problem (ii) corresponds to density of $\bar { p } _ { \mathrm { a u g } } ( \pmb { x } ^ { \prime } )$ outside of $p _ { \mathrm { d a t a } } ( \pmb { x } )$

 For VAEs, we refer to the difference between training and test set ELBO as the generalization gap  Therefore, $\mathcal { G } _ { \mathrm { g } } \geq 0$   A smaller $\mathcal { G } _ { \mathrm { g } }$ corresponds to a better generalization performance of a VAE Remark (Test data entropy can also affect the ELBO value)
\\\\
\textbf{MemoRAG:} ... ( $\mathcal G_\text{g} \geq 0$ ) holds when player ( i ) maintains sufficient SCs and units, avoiding disbandment; ( $\mathcal G_\text{g} < 0$ ) occurs if player ( i ) has few SCs, disbands units, or is effectively blocked by player ( -i ).

\end{exampleblock}

\vspace{0.2cm}

\begin{exampleblock}[]
\textbf{Question:} What are the novel aspects of the proposed method compared to prior work such as Lee et al. (2019), and how do they contribute to the improvement in performance?
\\\\
\textbf{Answer:} ... The improvement in performance over ICT mainly comes from three factors: using MoCo to handle negatives, which allows scaling to a large number of negatives; the sampling procedure to generate pairs of (query, key); and using data from both CC-net and Wikipedia for training. The paper will be updated to highlight the origin of these gains. ...
\\\\
\textbf{Retrieve:} A downside of this approach is that it requires extremely large batch sizes to work well Chen et al  (2020), with Qu et al  (2021) reporting improvement in the context of information retrieval up to 8192 negatives  This method has been widely used to train information retrieval models with supervised data Chen et al  (2017b); Karpukhin et al  (2020) and was also considered when using ICT to pre-train retrievers by Lee et al

Hence, we believe that most of the improvement compared to the state-of-the-art retrievers can be attributed to our contrastive pre-training strategy Table 7: Impact of data augmentions  We report nDCG@10 without fine-tuning on MS MARCO

Third, when used as a pre-training method before fine-tuning on MS MARCO, our technique leads to strong performance on the BEIR benchmark  We perform ablations to motivate our design choices, and show that cropping works better than the inverse Cloze task  Finally we train a multilingual dense retriever with contrastive learning and show that it achieves state-of-the-art performance Code and pre-trained models are available here: https://github com/facebookresearch/contriever  2 Related work

It should be noted that our fine-tuning procedure on MS MARCO is simpler than for other retrievers, as we use a simple strategy for negative mining and do not use distillation  Our model would probably also benefits from improvements proposed by these retrievers, but this is beyond the scope of this paper Finally, we illustrate the benefit of our retriever compared to BM25 in a few-shot setting, where we have access to a small number of in-domain retrieval examples

Interestingly, as shown in Table 7, the random cropping strategy outperforms the inverse cloze task in our setting  We believe that random cropping, leading to the identical distributions of keys and queries, leads to more stable training with MoCo compared to ICT  This might explains part of the difference of performance between the two methods  We also investigate whether additional data perturbations, such as random word deletion or replacement, are beneficial for retrieval Training data

In this section, we briefly review relevant work in information retrieval, and application of machine learning to this problem  This is not an exhaustive review, and we refer the reader to Manning et al  (2008), Mitra et al  (2018) and Lin et al  (2020) for a more complete introduction to the field Term-frequency based information retrieval

The methods described in the previous paragraph were applied to re-rank documents, which were retrieved with a traditional IR system such as BM25  Gillick et al  (2018) first studied whether continuous retrievers, based on bi-encoder neural models, could be viable alternative to re-ranking  In the context of question answering, Karpukhin et al  (2020) introduced a dense passage retriever (DPR) based on the bi-encoder architecture

We report results, without fine-tuning on MS MARCO in Table 6  We observe that the difference of performance between the two methods is small, especially after fine-tuning on MS MARCO  We thus propose to use MoCo as our contrastive learning framework, since it scales to a larger number of negative examples without the need to increase the batch size Number of negative examples

Finally, we also consider additional data augmentations such as random word deletion, replacement or masking  We use these perturbations in addition to random cropping  3 1 3 Building large set of negative pairs An important aspect of contrastive learning is to sample a large set of negatives  Most standard frameworks differ from each other in terms of how the negatives are handled, and we briefly describe two of them, in-batch negative sampling and MoCo, that we use in this work

arXiv preprint arXiv:2002 08909, 2020  3, 5, 7 Kaiming He, Haoqi Fan, Yuxin Wu, Saining Xie, and Ross Girshick  Momentum contrast for unsupervised visual representation learning  In Proceedings of the IEEE/CVF Conference on Computer Vision and Pattern Recognition, pp  9729–9738, 2020
\\\\
\textbf{MemoRAG:} The proposed method introduces a DRL-based heuristic that generates local moves via a deep policy network, learns longer-sighted search policies, utilizes a GNN-based representation for JSSP, and employs a message-passing mechanism for efficient solution evaluation, leading to improved performance over prior methods.
\end{exampleblock}

\vspace{0.2cm}

\begin{exampleblock}[]
\textbf{Question:} What is the purpose and methodology of the In-breadth Evolving analysis, and how does it demonstrate the diversity of the dataset compared to ShareGPT and Alpaca (Self-Instruct)?
\\\\
\textbf{Answer:} The Analysis of In-breadth Evolving section aims to enhance topic coverage, skill coverage, and overall dataset diversity. The methodology involves using BERT to encode instructions into 768-dimensional embeddings, applying t-SNE to reduce the embeddings to 2 dimensions, and using k-means clustering to partition instructions into 20 clusters for visualization. As shown in Figure 7, the data points of the dataset are more dispersed than those of ShareGPT and Alpaca (Self-Instruct), indicating better topic diversity.
\\\\
\textbf{Retrieve:} In-breadth Evolving aims to enhance topic coverage, skill coverage, and overall dataset diversity

{Here is instruction of Example N }    Rewritten Prompt: In-Breadth Evolving aims to enhance topic coverage, skill coverage, and overall dataset diversity  Open-domain instruction finetune datasets (e g , Alpaca, ShareGPT, etc ) are typically small in scale, lacking topic and skill diversity  To solve this problem, we designed a prompt to generate a completely new instruction based on the given instruction, requiring the new instruction to be more long-tailed

 As shown in the Figure 7, the data points of our dataset are more dispersed than ShareGPT and Alpaca (Self-Instruct), which indicates the better topic diversity in our instructions

The results indicate that (i) the ShareGPT is a better seed for evol-instruct than Alpaca, (ii) larger evolved data size can improve model capacity, and (iii) our proposed Evol-Instruct method is not dependent on ChatGPT, other strong open source model such as Llama-2 is also a good substitute for <table><tr><td>Model</td><td>Avg

, Llama-1 65B, Llama-2, Mistral-7B) indicate that our Evol-Instruct can be widely applied to various pre-trained models Analysis of In-depth Evolving  The Figure 5a and 5b presents an ablation study investigating the impact of the number of data evolution rounds  To study the depth of the evolving process, we use ChatGPT to judge the difficulty level of instruction  The used prompt please refer to Appendix E

, 2023a), and Tulu (Wang et al , 2023)  4 2 EXPERIMENT DETAIL To construct the dataset, we initialize it with the $5 2 k$ instruction dataset of Alpaca and iteratively perform $M$ evolutions, where $M = 4$   For each instruction in each round of evolution, we randomly select one evolving prompt from total six prompts (i e

Table 2: WizardLM with different data seed, data size, evol model, and base model size  Figure 5: The difficulty level between ShareGPT, Alpaca, and our four epochs of evolved instruction

In order to study the impact of different data seeds Evol models scale of evolved dataset pre-trained models on our proposed method we conducted the following experiments a) Using 70k ShareGPT as the seed data to obtain WizardLM-13b (ShareGPT Seed) b) Using LlaMA-2-70B-Chat to replace ChatGPT as the evolutionary execution model to obtain WizardLM-13b (LlaMA-2-70B-Chat Evol)

 To examine (qualitative analysis) the breadth (diversity) of different dataset, we firstly use BERT to encode each instruction and get its embedding with 768 dimensions, then use a dimension reduction algorithm named t-SNE to reduce embedding dimension to 2, finally we apply a clustering algorithm $\mathbf { k }$ -means to partition the instructions of each dataset into 20 clusters for an intuitive visualization

 Alpaca used a dataset of $5 0 \mathrm { k }$ instructions generated from a limited (e g , 175 samples) seed set of manually-written instructions  Our work is different from InstructGPT and Vicuna in that we use AI-generated data for instruction fine-tuning  Unlike Alpaca’s self-instruct Wang et al  (2022a) generation method, Evol-Instruct can control the difficulty and complexity level of the generated instructions
\\\\
\textbf{MemoRAG:} The purpose of the In-breadth Evolving analysis is to enhance topic coverage, skill coverage, and overall dataset diversity. The methodology involves encoding each instruction using BERT to obtain 768-dimensional embeddings, reducing the dimensions to 2 using the t-SNE algorithm, and then applying the $k$-means clustering algorithm to partition the instructions into 20 clusters for visualization. The analysis demonstrates the diversity of the dataset by showing that the data points in the cluster scatter plot are more dispersed compared to ShareGPT and Alpaca (Self-Instruct), indicating better topic diversity in the instructions.
\end{exampleblock}

\section{Benchmark Details} \label{sec:benchmark_details}

\subsection{Practical Implications} \label{sec:practical_implications}

Beyond benchmark construction and model comparison, the benchmark also provides practical value for real-world paper-centric workflows. In particular, it can support practitioners in four aspects.

\textbf{Research paper comprehension.} The benchmark evaluates whether a model can faithfully understand full papers, including core concepts, methods, and experimental findings. This makes it useful for comparing models in workflows such as literature review, method comparison, and experiment interpretation, where reliable paper understanding is critical.

\textbf{Long-context evaluation.} Because many questions require evidence integration across an entire paper, the benchmark measures document-level understanding rather than local text matching. Practitioners can therefore use it to assess whether longer context windows or long-context architectures lead to meaningful gains in holistic reasoning over papers.

\textbf{Multimodal reasoning.} The benchmark includes both text inputs and page-level images, requiring models to interpret figures, tables, and diagrams in their original PDF layout. This reflects realistic scientific reading conditions and helps practitioners evaluate whether a model can effectively use visual evidence in addition to text.

\textbf{RAG system diagnosis.} Since the questions emphasize reasoning and synthesis over relevant paper content, the benchmark can also be used to diagnose retrieval-augmented generation systems. In particular, it helps practitioners compare retrieval, chunking, and evidence-fusion strategies, and identify which components improve paper-level comprehension rather than only snippet-level retrieval accuracy.

\subsection{Comparison with Other Paper-Related Benchmarks} \label{sec:paper-related_benchmarks}

Several prior benchmarks are also related to research papers, but they differ from RPC-Bench in task formulation, input setting, and evaluation.

SciFact~\citep{wadden2020fact} focuses on claim verification using scientific claims and abstract-level evidence. QASPER~\citep{dasigi2021dataset} mainly targets information-seeking question answering, with questions constructed from limited paper context such as the title and abstract. ScholarQABench~\citep{asai2024openscholar} emphasizes literature-search and synthesis settings over a literature datastore. ResearchQA~\citep{yifei2025researchqa} is designed for topic-level evaluation based on survey papers, rather than fine-grained understanding of individual research papers. 

In contrast, RPC-Bench is content-driven and organized around the research workflow of a full paper. It evaluates whether models can answer what, how, and why questions about concepts, methods, and experiments using evidence distributed across sections. RPC-Bench also supports both textual and visual inputs, enabling evaluation under more realistic paper-reading conditions. In evaluation, RPC-Bench adopts multi-judge scoring over correctness, completeness, and conciseness, whereas these benchmarks mainly rely on task-specific automatic metrics or literature-search-oriented criteria.

\subsection{Quality‑Based Filtering Criteria} \label{sec:data_filtering}

We perform quality-based filtering at two distinct stages (comment–response pairs and QA items) to remove low-quality items that cannot be reliably answered using only the paper's content. The filtering criteria are:

\begin{itemize}
    \item Temporary or editorial issue: corrections of grammar/spelling errors (e.g., "We corrected 'benchamrks' to 'benchmarks'"), figure color/font adjustments, formatting changes or adding references, open-sourcing code/data (e.g., "Added reference to Smith et al."), where the response merely acknowledges the fix without academic substance.
    \item External resource dependency: responses whose validity depends on external materials not contained in the paper (e.g., "More cases: https://..."), or indirect or evasive replies (e.g., "See Section X").
    \item Non-substantive commitments: promises of future additions (e.g., "We will add a limitations section", "Will address in future work") without providing specific details or a concrete resolution in the current submission.
\end{itemize}

\subsection{Scale and Scalability} \label{sec:scale_scalability}

\begin{table*}[h]
\small
    \newcolumntype{C}{>{\centering\arraybackslash}c}
    \centering
    \renewcommand\arraystretch{1.0}
    \begin{tabular}{c|@{~ }*{1}{CCCC}}
            \toprule[1.2pt]
        {Benchmarks}  &  {Papers}  & {QA} & {Question Context} & {Annotator} \\
        \midrule
        QASPER~\citep{dasigi2021dataset} & 1,585 & 5,049 & Abstract & Practitioners  \\
        BioASQ~\citep{krithara2023bioasq} & - & 4,615 & - & Experts \\
        QASA~\citep{lee2023qasa} & 113 & 1,798 & Full Paper & Practitioners  \\
        PeerQA~\citep{baumgartner2025peerqa} & 208 & 579 & Full Paper & Experts  \\
        RPC-Bench (dev+test) & 1,050 & 15,682 & Full Paper & Reviewers, Authors, LLMs, Skilled Annotators  \\
        \bottomrule[1.2pt]
    \end{tabular}
    \caption{Comparison of RPC-Bench with prior paper-centric QA benchmarks.}
    \label{table:comparison_scale}
\end{table*}

\paragraph{Scale.} The dev and test sets contain 15,682 human-validated QA pairs from 1,050 papers, which is comparable to or larger than several prior paper-centric human-annotated benchmarks (Table \ref{table:comparison_scale}). This scale is sufficient to support fine-grained evaluation and reliable model comparison. In addition, the training set includes 3,100 papers and 45,651 QA pairs and is released as an auxiliary resource for incremental use, allowing researchers to flexibly select or refine subsets without additional data construction.

\paragraph{Scalability.} Our framework separates taxonomy from data construction and evaluation. The taxonomy is grounded in a research-flow perspective (e.g., motivation, methodology, and experiments) rather than a fixed paper structure. This allows lightweight adaptation to domains whose papers differ in format or emphasis. By contrast, the data construction pipeline, annotation protocol, and evaluation methodology are reusable across domains. 
The framework also supports extension beyond AI/ML when suitable review-response style resources, such as open peer reviews or comment-reply articles, are available.

\subsection{Domain bias and distributions} \label{sec:domain_bias}

We adopted three measures to ensure balanced question difficulty and topic coverage:

\begin{itemize}
\item Papers with more than 50 citations were chosen to ensure strong academic impact. These highly cited papers are also the ones people typically want to explore in greater depth. To avoid potential bias introduced by citation count and acceptance status, we also included highly cited rejected papers and randomly sampled rejected papers. This expands coverage to emerging and less‑mainstream areas, reducing over‑representation of popular topics.
\item \benchmark spans top computer science conferences from 2013 to 2024 across multiple subfields, capturing trends and methodological diversity. Domain distribution includes ML Theory (24.80\%), Computer Vision (16.87\%), NLP (15.17\%), Reinforcement Learning (11.42\%), Optimization (7.97\%), Generative Models (6.69\%), Graph ML (6.36\%), AI for Science (3.23\%), and other areas, yielding balanced topic distribution.
\item Following the taxonomy, each question is categorized from lower‑complexity “what” types  
to higher‑complexity “how” 
and “why” types 
, covering theory (38.52\%)  through applications (61.48\%). As shown in Fig. \ref{fig:taxonomy}, the category distribution is well balanced, indicating no significant bias in difficulty during dataset construction.
\end{itemize}

\subsection{Potential Bias and Circularity from LLM Use} \label{sec:bias_circularity}

We mitigate potential bias and circularity from LLM use in benchmark construction and evaluation by restricting LLMs to auxiliary roles, preserving human supervision for the dev/test sets, and reducing reliance on any single model. 

\begin{table}[h]
\small
\centering
\renewcommand\arraystretch{1.0}
\begin{tabular}{lccc}
\toprule[1.2pt]
 & GLM & DeepSeek & Human \\
\midrule
Question & 29.20 & 52.15 & 18.65 \\
Answer   & 11.99 & 22.48 & 65.53 \\
\bottomrule[1.2pt]
\end{tabular}
\caption{Source distribution (\%) of question and answer revisions in the dev/test sets.}
\label{tab:qa_source_distribution}
\end{table}

The benchmark is derived from authentic review-rebuttal data, while LLMs are used only for decomposition, rewriting, and filtering under constraints that preserve the original meaning and avoid introducing new information.
In the dev/test sets, questions are mainly LLM-assisted rewrites with human review, whereas answers are primarily rewritten by annotators to ensure factual correctness and semantic fidelity (Table \ref{tab:qa_source_distribution}).
For the training set, we release filtered rewrites from multiple models rather than a single LLM-dependent version, enabling flexible bias control and robustness studies. For evaluation, we select the two judges with the highest agreement with human assessments and average their scores, reducing model-specific bias and mitigating evaluation-side circularity.

\subsection{Discussion on Potential Data Contamination} \label{sec:data_contamination}

A potential concern is that some publicly available papers in our benchmark may have been seen during model pretraining. However, our benchmark primarily evaluates paper comprehension rather than memorization. Answering the questions requires identifying key evidence in long documents, integrating multi-modal information, and reasoning across sections. Simple recall of isolated content is generally insufficient. Similar to human researchers, partial prior exposure to a paper does not imply genuine understanding.

The empirical results also do not support a simple data-exposure explanation. If memorization were the main driver, performance would be expected to improve monotonically with model recency, as shown in Table \ref{table:additional_main}. Instead, even GPT-5 reaches only 68.20\% F1-like and 54.93\% Conciseness, and newer models do not consistently outperform earlier ones: GPT-5.2 declines overall, while Claude-4.5 and Gemini-3 improve in F1-like but reduce conciseness. This non-monotonic pattern suggests that the benchmark mainly tests information integration and reasoning rather than direct recall.

To further mitigate contamination, we adopt three measures: 
(1) Temporal split. 2024 papers are used exclusively for testing, with 2022–2023 reserved for validation, minimizing overlap with pretraining data;
(2) QA rewriting. Dev/test QA pairs are decomposed, rewritten by LLMs, and manually verified to substantially diverge from the original review–rebuttal text and reducing the chance of memorized QA pairs;
and (3) Controlled evaluation. All models are evaluated under identical settings, enabling fair horizontal comparison among contemporaneous models and mitigating residual contamination effects.

\subsection{Motivation of "Informativeness" Metric} \label{sec:motivation_info}

The Informativeness metric integrates correctness, completeness, and brevity, offering a balanced view of paper comprehension quality. Among these, conciseness is not merely a stylistic preference; it is a crucial quality factor that reduces redundancy and semantic noise. Separate dimension assessment may overlook interaction effects, whereas informativeness captures overall performance in a holistic way that prevents inflated scores from verbose outputs. For example, assuming a ground-truth answer is “A, B, C, D, E,” a candidate answer is “A, A, A, B, B, B, C, C, C, D, D, D.” This candidate answer achieves 100\% precision, 80\% recall, and 88.89\% F1. However, its output is verbose. This answer should be penalized in terms of conciseness.

\subsection{Annotation Details} \label{sec:detail_annotation}

All participants were provided with written instructions prior to beginning the task. The instructions explicitly described the task objectives, the expected time commitment, and concrete examples of acceptable and unacceptable annotations. The instructions stated that participation was voluntary and that participants could stop at any time without penalty. 

We recruited four annotators with formal training in computer science, all holding a Master’s degree or higher. Annotators were compensated at a rate of 1\$ per successfully annotated instance. Based on observed annotation speed of approximately 5–6 minutes per question (about 80 QA pairs per day), this corresponds to an effective hourly wage that is competitive with and above typical research assistant rates in their local contexts.
All annotators were all adults currently residing in Asia. No protected information (e.g., sexual orientation or political views under GDPR) were collected or included in the dataset.

Informed consent was obtained from all participants prior to data collection. Before accessing the task, participants were required to read a consent statement explaining the purpose of the study, the nature of the data being collected, and how the data would be stored, processed, and used in academic publications. Participants were informed that their anonymized annotations would be used for research purposes and could be released as part of a publicly available dataset. Proceeding with the task was taken as an explicit indication of consent.

The data collection protocol was reviewed by an institutional ethics review process and was determined to be exempt from full review under applicable regulations, as the study involved minimal risk, anonymous data collection, and no collection of personal identifying information.

The full set of prompts used in this stage are provided in Appendix \ref{sec:prompt_decompose} and Appendix \ref{sec:prompt_conversion}.

Using the annotation platform (Appendix \ref{sec:screenshot_annotation}), annotators examined each segmented review–rebuttal pair and chose the better output between GLM-4-Plus and DeepSeek-V3, while verifying taxonomy labels. If both outputs were inadequate, they rewrote the pair manually and assigned the correct category. To reduce bias, model identities were anonymized as Model1 and Model2, with randomized ordering. Annotators could discard low-quality pairs or generate multiple sub-questions from a single pair, provided each addressed a distinct aspect. 

The review platform (Appendix \ref{sec:screenshot_review}) displayed both original and rewritten content, allowing reviewers to approve or reject entries. Rejections required specific feedback to guide further revisions. 

\subsection{Annotation Platform} \label{sec:screenshot_annotation}

\begin{figure*}[h]
\centering
\includegraphics[width=0.95\linewidth]{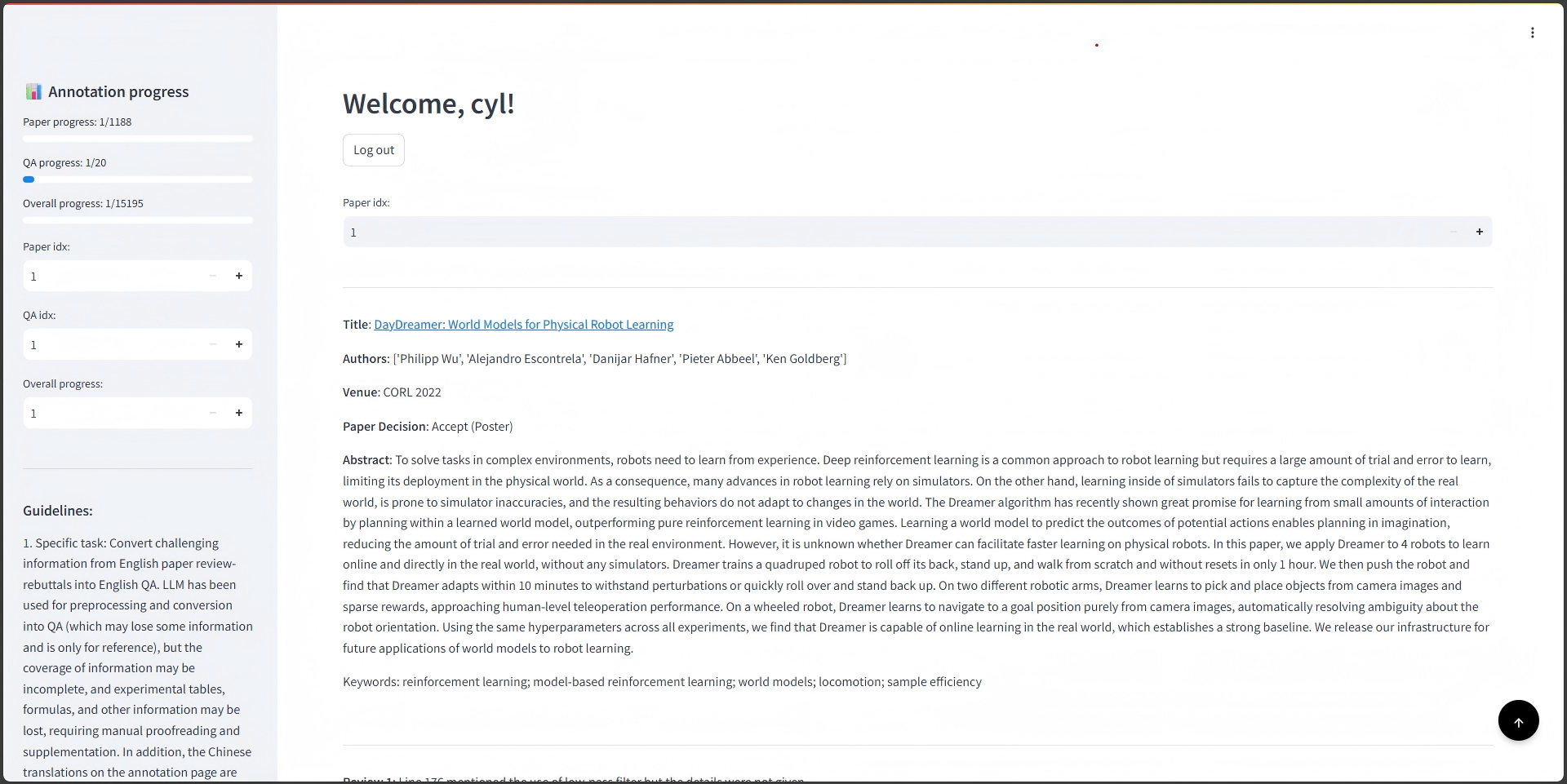}
\caption{Screenshot of the Annotation Interface 1}
\end{figure*}

\begin{figure*}[h]
\centering
\includegraphics[width=0.95\linewidth]{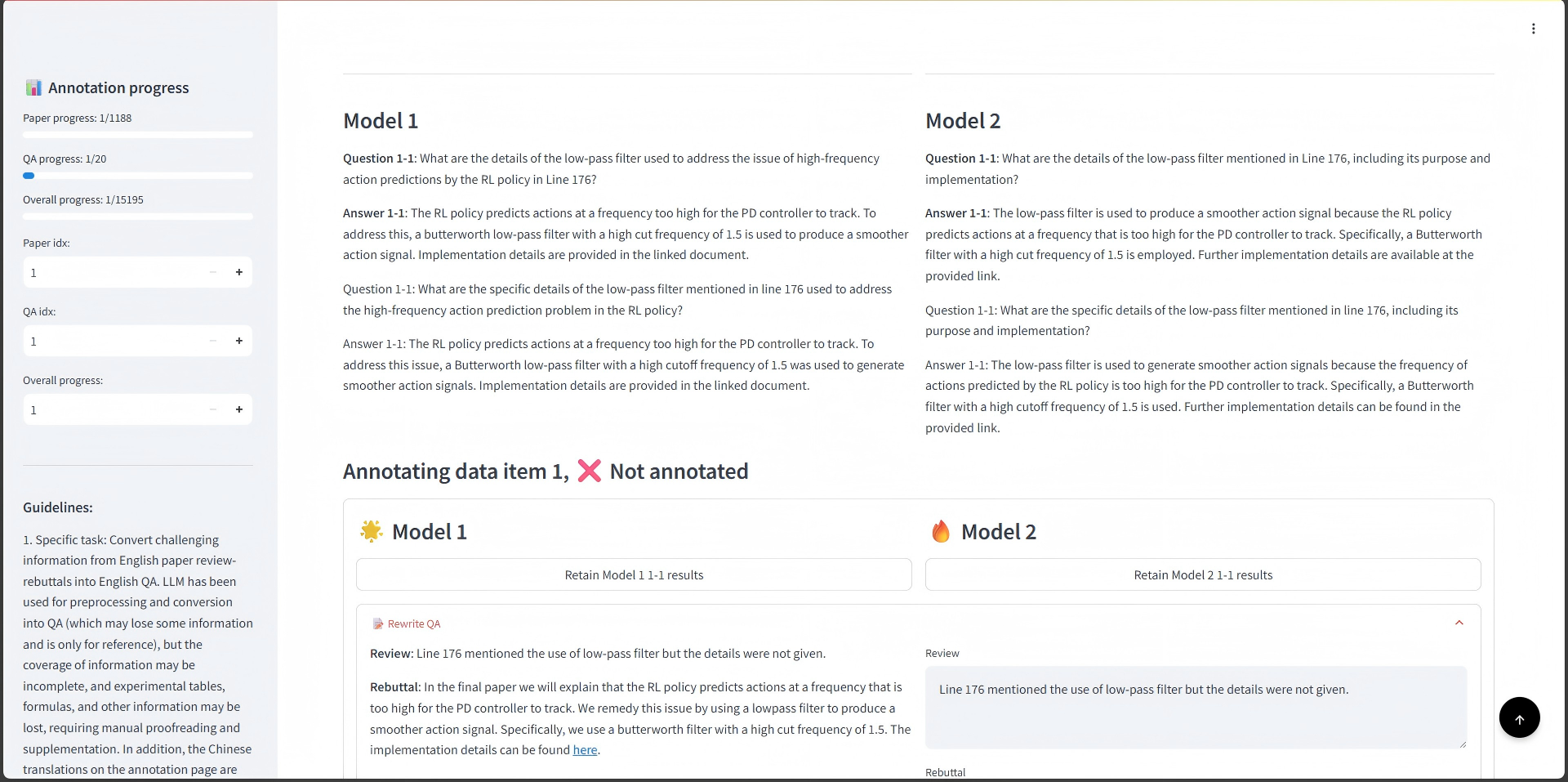}
\caption{Screenshot of the Annotation Interface 2}
\end{figure*}

\begin{figure*}[h]
\centering
\includegraphics[width=0.95\linewidth]{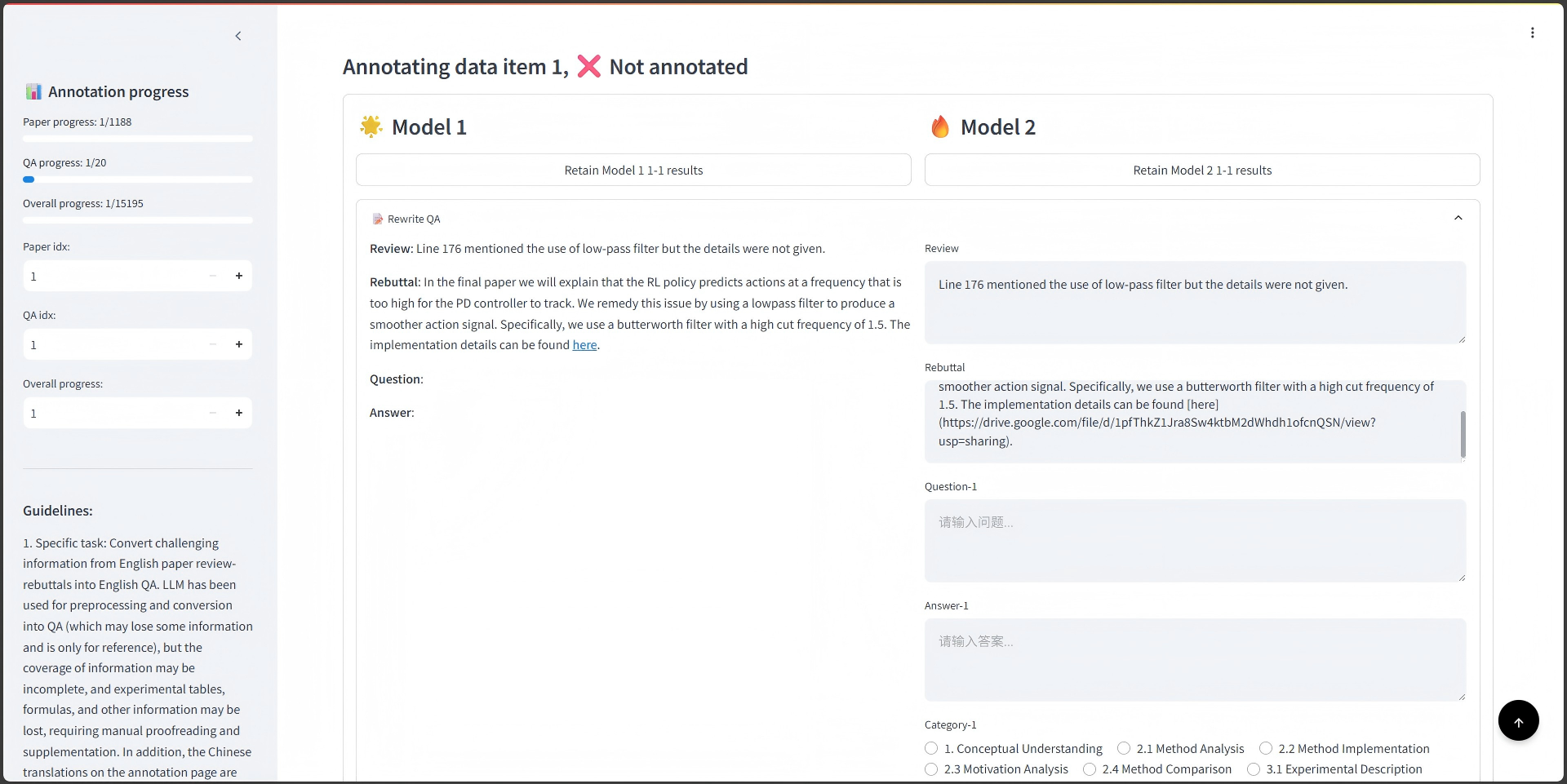}
\caption{Screenshot of the Annotation Interface 3}
\end{figure*}

\subsection{Review Platform} \label{sec:screenshot_review}

\begin{figure*}[h]
\centering
\includegraphics[width=0.95\linewidth]{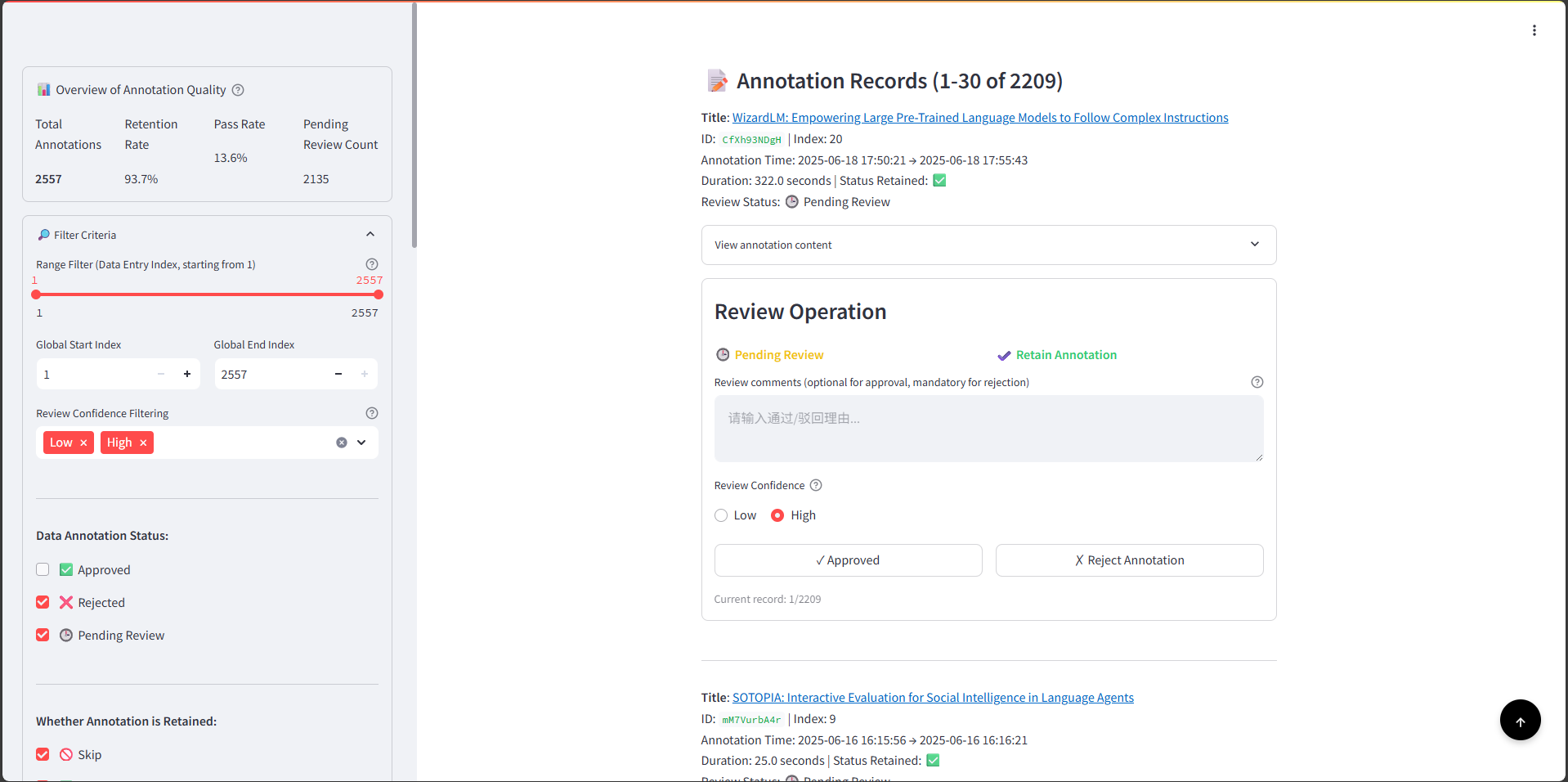}
\caption{Screenshot of the Review Interface 1}
\end{figure*}

\begin{figure*}[h]
\centering
\includegraphics[width=0.95\linewidth]{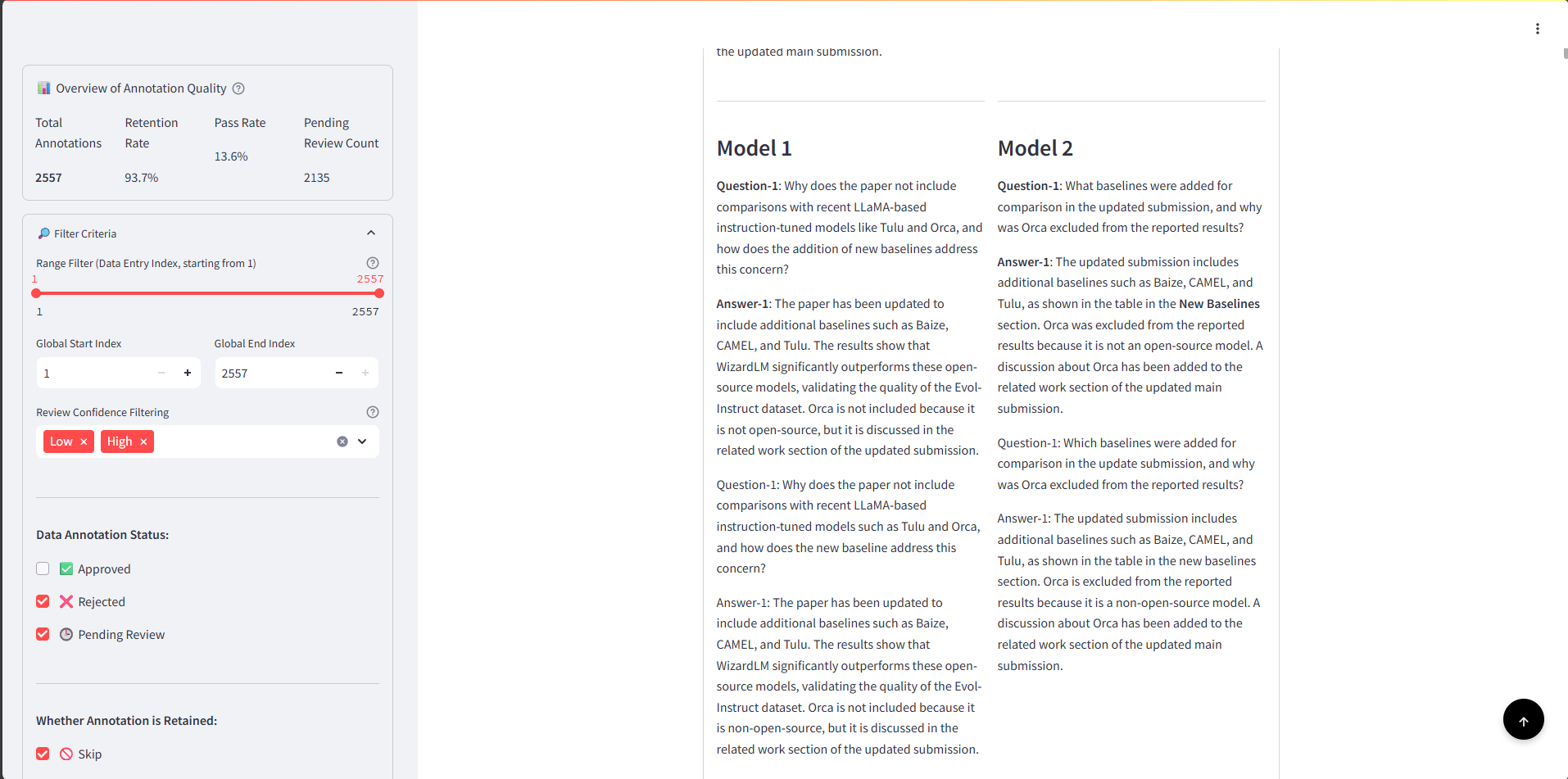}
\caption{Screenshot of the Review Interface 2}
\end{figure*}

\begin{figure*}[h]
\centering
\includegraphics[width=0.95\linewidth]{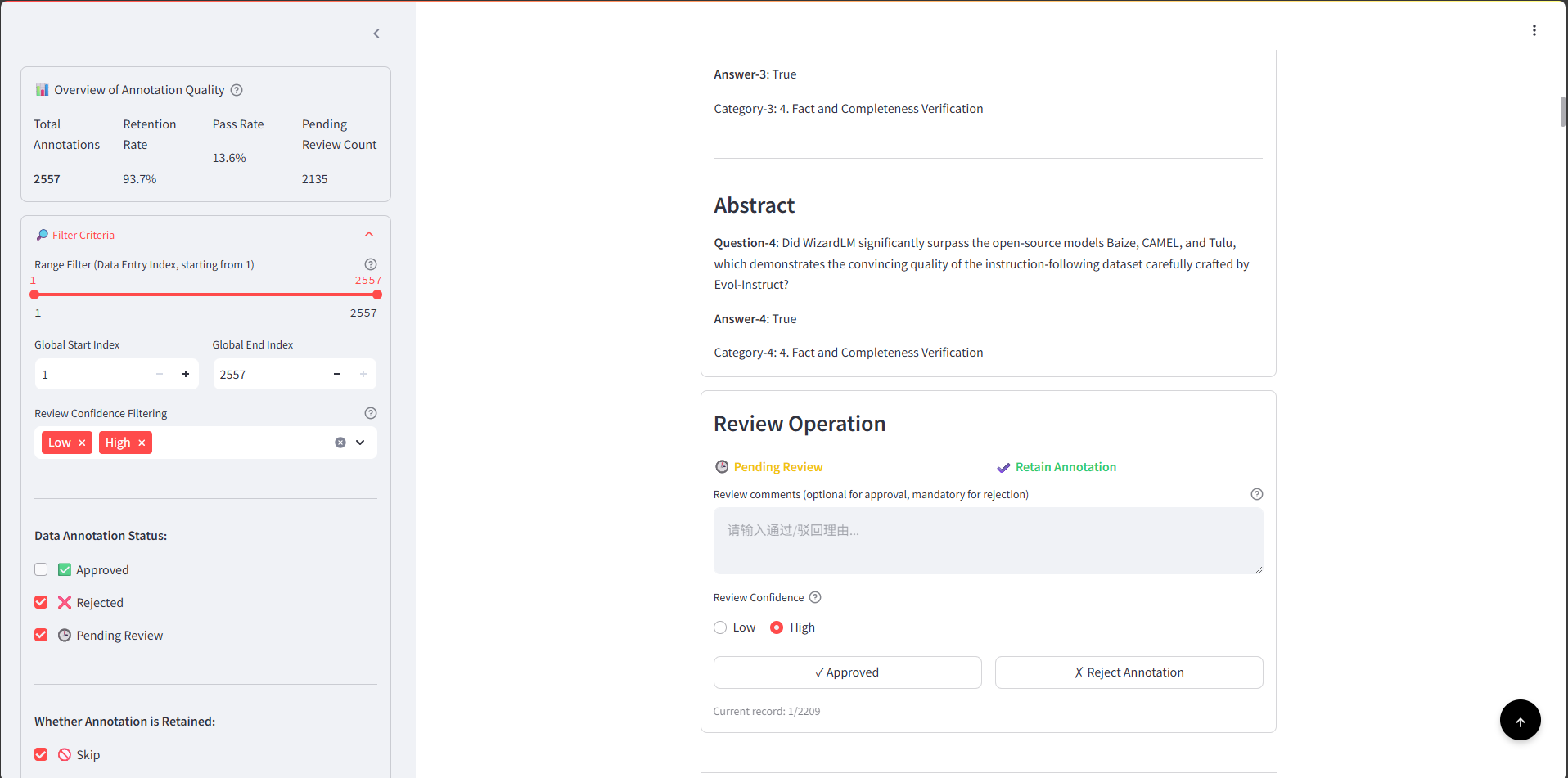}
\caption{Screenshot of the Review Interface 3}
\end{figure*}

\subsection{Decompose Prompt} \label{sec:prompt_decompose}

\begin{lstlisting}[style=prompt]
You are an excellent reviewer of papers. You are tasked with extracting QA pairs from the "review", "rebuttal" and "extra_rebuttal" sections of a conference paper submission. This process includes identifying "review" provided by reviewers and pairing them with the corresponding answers authored by the paper's authors, utilizing content from both the "rebuttal" and any relevant "extra_rebuttal" sections. 
Your goals are: Extract and classify the QA pairs. Ensure that references and citations in the rebuttal are preserved in their original format within the answers, maintaining the academic rigor and clarity. Determine whether each question-answer pair is 'multimodal-related,' a broad concept that includes questions explicitly about the figures and tables in the paper or questions that can only be answered by referring to the contents of these figures and tables.

Input Structure:

review: Concatenation of all reviews, including multifaceted evaluations of the paper and any responses or questions directed at the authors' rebuttal.
rebuttal: The content in the rebuttal is a concatenation of the answers to all the review questions.
extra_rebuttal: Additional content from the authors that may cover the current questions.

Output Requirements:

For each QA pair, output in the following JSON format:
[
    {
        "question": "extracted question text here",
        "answer": "corresponding answer text here",
        "is_multimodal_related": true or false
    },
    ...
]

Guidelines:

1. Split combined questions into finer sub-questions for clarity but merge them if they cannot stand alone meaningfully.
2. Ensure the completeness and consistency of the extracted QA pairs.
3. Use content from the extra_rebuttal to enhance or clarify answers when applicable and relevant to the question.
4. Ensure that the rebuttal content is fully utilized in the answers, forming comprehensive and clear QA pairs that correspond to the questions posed.
5. Use your judgment to label each QA pair as 'multimodal-related' if it either explicitly poses questions about the figures and tables in the paper or implicitly requires the content of these figures and tables to answer the question.
6. The answers should be as comprehensive as possible, retaining any relevant content such as "references" that can assist in addressing the questions.
7. Use the original content from the review, rebuttal, and extra_rebuttal to construct the QA pairs, avoiding unnecessary modifications to the original text.

Input:
review: It is novel enough to combine the advantages of two famous models (Transformer, RNN). Also, the combining method looks applicable to a variety of scenarios. The experimental results are impressive, showing superior performance to previous Transformer.

I think the draft would become better if there is a more complete explanation and figures about the self-attention with recurrence (RSA) operation.

I think the novelty of this draft is enough for the publication and the experimental results are impressive. English is good enough as well. I recommend weak accept for the draft.

rebuttal: Thanks for your encouraging words and constructive comments. We sincerely appreciate your time in reading the paper, and our point-to-point responses to your comments are given below.

> I think the draft would become better if there is a more complete explanation and figures about the self-attention with recurrence (RSA) operation.

Thank you for this instructive comment. Following your suggestions, we have provided a graphical illustration of a single headed RSA module in Figure 1 (d) on Page 2, and a more detailed explanation about the operation of RSA has been given in the paragraph of "Operation of multihead RSA modules" on Page 5.       

In the meanwhile, we have also reorganized the whole Section 3 to better explain the proposed RSA. Specifically,
For a single head RSA, we have devoted a paragraph right after equation (4) to detail the different types of REMs i.e. $\mathbf{P}$ in the paper.

For your easy reference, we have listed the multihead RSA operation below:

        Procedure for the Multihead RSA
                - Choose masked or unmasked REMs according to the nature of the task.
                - Select the hyperparameters including the dilating factor $d$ and the numbers of the six types of REMs $(k_1,\dots,k_6)$.
                - For each head, apply equation (4) with a different REM.
                - Apply a linear layer to combine the output from all heads, and perform layer-normalization and dropout.

extra_rebuttal: We will make the following revisions to the paper:

1. Block-Recurrent Transformer (BRT) [1] has been adopted as another baseline model for the NLP experiment in Section 4.3, and its results are presented as follows.

|                             | BRT        | RSA-BRT    |
| --------------------------- | ---------- | ---------- |
| Enwik8                      | 1.0746     | **1.0683** |
| Text8                       | 1.1652 | **1.1625** |
| WikiText-103                | 23.758     | **23.639** |
| # Averaged Params added (%) |            | 8.68E-05   |

It can be seen that RSA-BRT exceeds the baseline BRT's performance on all datasets.

**The results of this table will be used to fill in the blanks in Table 3 (b) of the paper.**

2. Two additional experiments for Section 4.4 have been conducted during the second discussion phase, which are detailed in the responses to Reviewers mvWh and Zrmk.

(1) A scaling experiment is conducted for RSA-BRT v/s BRT on Enwik8 dataset. The results are shown as follows.

| #  layers           | 8          |           | 10         |           | 12         |           | 14         |           |
| ------------------- | ---------- | --------- | ---------- | --------- | ---------- | --------- | ---------- | --------- |
|                     | Params     | BPC       | Params     | BPC       | Params     | BPC       | Params     | BPC       |
| BRT                 | 35,080,908 | 1.127     | 41,905,868 | 1.106     | 48,730,828 | 1.098     | 55,555,788 | 1.079     |
| RSA-BRT             | 35,080,943 | **1.120** | 41,905,913 | **1.104** | 48,730,883 | **1.092** | 55,555,853 | **1.072** |
| Increase in #Params | 35         |           | 45         |           | 55         |           | 65         |           |

It can be seen that, with only less than 100 new parameters, RSA-BRT can achieve some improvement over the baseline BRT. More importantly, the advantage can be consistently observed for all model sizes.

(2) Another scaling experiment is conducted for RSA-XL against TL-XL on Text8 dataset, where REM is replaced by a learnable Toeplitz matrix in the latter model. The results are shown as follows.

| #  layers           | 8          |           | 10         |           | 12         |           | 14         |           |
| ------------------- | ---------- | --------- | ---------- | --------- | ---------- | --------- | ---------- | --------- |
|                     | Params     | BPC       | Params     | BPC       | Params     | BPC       | Params     | BPC       |
| TL-XL               | 34,180,645 | 1.193     | 41,013,799 | 1.188     | 47,846,953 | 1.183     | 54,680,107 | 1.178     |
| RSA-XL              | 34,139,725 | **1.181** | 40,964,695 | **1.170** | 47,789,665 | **1.164** | 54,614,635 | **1.160** |
| Decrease in #Params | 40,920     |           | 49,104     |           | 57,288     |           | 65,472     |           |

From the above table, it can be seen that the newly added TL-XL also performs worse than the RSA-XL of a similar model size, indicating parameter redundancy. In other words, RSA-XL enjoys a much better parameter-efficiency.

**These two experiments will be further included into Section 4.4 of the paper.**

Reference

[1] Hutchins, D., Schlag, I., Wu, Y., Dyer, E., and Neyshabur, B. (2022). Block-recurrent transformers. In Advances in Neural Information Processing Systems.

Output:
[
    {
        "question": "I think the draft would become better if there is a more complete explanation and figures about the self-attention with recurrence (RSA) operation.",
        "answer": "Thank you for this instructive comment. Following your suggestions, we have provided a graphical illustration of a single headed RSA module in Figure 1 (d) on Page 2, and a more detailed explanation about the operation of RSA has been given in the paragraph of 'Operation of multihead RSA modules' on Page 5. In the meanwhile, we have also reorganized the whole Section 3 to better explain the proposed RSA. Specifically, for a single head RSA, we have devoted a paragraph right after equation (4) to detail the different types of REMs i.e. $\\mathbf{P}$ in the paper. For your easy reference, we have listed the multihead RSA operation below: Procedure for the Multihead RSA - Choose masked or unmasked REMs according to the nature of the task. - Select the hyperparameters including the dilating factor $d$ and the numbers of the six types of REMs $(k_1,\\dots,k_6)$. - For each head, apply equation (4) with a different REM. - Apply a linear layer to combine the output from all heads, and perform layer-normalization and dropout.",
        "is_multimodal_related": true
    }
]

Input:
review: I would like to request further clarification regarding your paper after carefully reading it. Firstly, I would like to express my sincere appreciation for the captivating nature of your work and the clarity with which it is presented. Congratulations for the acceptance of your paper into the top 5% category.

In Section 4.3, I noticed the utilization of Transformer-XL with 14 layers, resulting in a notable achievement of 1.074 on the Enwik8 dataset. However, upon referencing the Transformer-XL paper, it became apparent that they reported lower bpc values, specifically 1.06 with 12 layers, 1.03 bpc with 18 layers, and an impressive 0.99 bpc with 24 layers.

To enhance my understanding, I kindly request your insights regarding the decision to opt for 14 layers instead and the possible reasons behind the relatively higher bpc despite employing deeper layers. Additionally, I would greatly appreciate any additional details or insights you can provide to address these inquiries.

Thank you in advance for your time and consideration. Your input will greatly contribute to my comprehension of your valuable research. Once again, congratulations on the successful publication of your paper.
rebuttal: Hi Lokesh, thanks for the question!

The observed difference between the reported bits per character (bpc) for Enwik8 in Section 4.3 of our paper and the original Transformer-XL paper can be attributed to our decision to utilize Nvidia's implemented Transformer-XL (https://catalog.ngc.nvidia.com/orgs/nvidia/resources/transformerxl_for_pytorch) rather than the official repository. We chose the Nvidia version due to its enhanced user-friendliness and comprehensive multi-card support.

However, it is important to note that the reproduction by Nvidia resulted in slightly worse bpc for Enwik8 compared to the figures reported in the original paper. Specifically, the bpc for Enwik8 with a 12-layer Transformer-XL exceeded the previously reported value of 1.06. This discrepancy could be attributed to variations in the implementation and environment between Nvidia's version and the official repository.

Furthermore, from an intuitive perspective, when a model is overparameterized, the proposed RSA may exhibit better generalization ability, as illustrated in Figure 1. In order to emphasize the benefits of the proposed RSA, we employed a slightly larger model. Unfortunately, due to limited resources, we were unable to conduct further experiments using a 24-layer XL model.

While acknowledging these limitations, we believe that the use of Nvidia's implementation, combined with our modifications, provides valuable insights and supports our argument. The comparison between the modified models, despite the slight deviations, offers meaningful observations regarding the potential advantages of the proposed RSA.

extra_rebuttal:
Output:
[
    {
        "question": "In Section 4.3, I noticed the utilization of Transformer-XL with 14 layers, resulting in a notable achievement of 1.074 on the Enwik8 dataset. However, upon referencing the Transformer-XL paper, it became apparent that they reported lower bpc values, specifically 1.06 with 12 layers, 1.03 bpc with 18 layers, and an impressive 0.99 bpc with 24 layers. To enhance my understanding, I kindly request your insights regarding the decision to opt for 14 layers instead and the possible reasons behind the relatively higher bpc despite employing deeper layers.",
        "answer": "The observed difference between the reported bits per character (bpc) for Enwik8 in Section 4.3 of our paper and the original Transformer-XL paper can be attributed to our decision to utilize Nvidia's implemented Transformer-XL (https://catalog.ngc.nvidia.com/orgs/nvidia/resources/transformerxl_for_pytorch) rather than the official repository. We chose the Nvidia version due to its enhanced user-friendliness and comprehensive multi-card support. However, it is important to note that the reproduction by Nvidia resulted in slightly worse bpc for Enwik8 compared to the figures reported in the original paper. Specifically, the bpc for Enwik8 with a 12-layer Transformer-XL exceeded the previously reported value of 1.06. This discrepancy could be attributed to variations in the implementation and environment between Nvidia's version and the official repository. Furthermore, from an intuitive perspective, when a model is overparameterized, the proposed RSA may exhibit better generalization ability, as illustrated in Figure 1. In order to emphasize the benefits of the proposed RSA, we employed a slightly larger model. Unfortunately, due to limited resources, we were unable to conduct further experiments using a 24-layer XL model. While acknowledging these limitations, we believe that the use of Nvidia's implementation, combined with our modifications, provides valuable insights and supports our argument. The comparison between the modified models, despite the slight deviations, offers meaningful observations regarding the potential advantages of the proposed RSA.",
        "is_multimodal_related": true
    }
]

Input:
review: -       The idea of utilizing dataset exchangeability to identify test set contamination is novel and interesting.
-       The proposed sharded likelihood comparison test addresses the tradeoff between statistical power and computational requirements of the permutation test, which is promising. The sharded rank comparison test also provides (asymptotic) guarantees on false positive rates.
-       Experimental results are promising. A GPT-2 model is trained from scratch on standard pretraining data and known test sets to verify the efficiency of the proposed method in identifying test set contamination. The method is also tested with an existing model, LLaMA2, on the MMLU dataset, showing general agreement with the contamination study results.   
-       Although a more efficient sharded rank comparison test is proposed, the computational complexity is still considerable. For example, testing 49 files using 1000 permutations per shard can take 12 hours for LLaMA2.
-       There is no comparison with other baseline methods.
-       The method relies on a strong assumption of data exchangeability, which may not hold in real-world datasets.     
If a dataset is not exchangeable, how effective is the method?

rebuttal: Thank you for your thorough review and valuable feedback on our work.

We'd like to address the concern regarding the computational complexity of our test. It's important to note that the test is a one-time process for any given model and dataset; once the p-values are computed, there is no need for recalculation. Our findings indicate that a number of permutations beyond 30-50 per shard offers diminishing returns, as shown in Figure 3 (right).

Furthermore, the test's design allows for easy parallelization. Each shard permutation can be evaluated independently, enabling the use of inexpensive commodity hardware to run the test significantly faster.

Regarding the assumption of data exchangeability, this is a strictly weaker condition than the commonly held assumption of independent and identically distributed (I.I.D.) data in machine learning. Most datasets satisfy this assumption to some extent.

We acknowledge the validity of our test hinges on data exchangeability. However, depending on the source of non-exchangeability, it is often the case that a dataset can be altered slightly so that our test is still valid. For example, a common source of non-exchangeability is the presence of ascending IDs (e.g., as in SQuAD and HumanEval). We can adjust the data-by either removing these IDs or permuting the examples while keeping IDs constant-to retain the test's applicability. This is discussed in more detail in the revised paper. 

Finally, we appreciate your suggestion to include baseline comparisons. We provide a comparison against a contamination detection method called Min-K% Prob, a state of the art heuristic method for contamination detection in language models proposed contemporaneous to our work by Shi et. al. (2023).

We find that our method matches or exceeds the performance of this state of the art heuristic method. Please see the table in the top-level comment for numbers.

extra_rebuttal: We are sincerely grateful to the reviewers for dedicating their time and effort to review our work, and we appreciate the recognition of the novelty of using exchangeability for contamination detection and the significance of our contribution given the discourse surrounding contamination in the field. We address each reviewer's comments in detail below. We have made numerous updates to the submission, most notably with the results of our test on four popular open models and eight commonly used benchmarks.

One question shared by multiple reviewers is regarding the exact notion of contamination we consider in this work. Rather than consider a definition based on heuristics like n-gram overlap, we consider contamination detection as the problem of detecting statistical dependence between the test data and model parameters. Within this setting, our work shows that it is possible to provide provable guarantees of contamination in the case of verbatim contamination, where the full test set (with examples and labels) is embedded in the pretraining data.

To illustrate the relevance of this setting, we note that a search of The Pile, a large open-source language modeling dataset, yielded numerous instances of small real-world datasets embedded with examples appearing in-order. As one example, the following is an excerpt from a dataset for an annotation tool made by Explosion, the creators of spaCy, a popular natural language processing framework, found in The Pile:

```
{"text":"Uber\u2019s Lesson: Silicon Valley\u2019s Start-Up Machine Needs Fixing","meta":{"source":"The New York Times"}}
{"text":"Pearl Automation, Founded by Apple Veterans, Shuts Down","meta":{"source":"The New York Times"}}
{"text":"How Silicon Valley Pushed Coding Into American Classrooms","meta":{"source":"The New York Times"}}

Source: https://github.com/explosion/prodigy-recipes/tree/fc06f6a6d93bc477e98cf0d8357c39322e4f5a6a
```

What our work shows is that by exploiting exchangeability in this setting, we are able to provide guarantees on the false positive rate of our test.

Multiple reviewers indicated the desire for a comparison against a baseline method. While no other existing work is comparable in the sense that it provides a statistical proof of contamination like ours, we provide a comparison against a state of the art heuristic method for contamination detection called Min-K% Prob, proposed by Shi et. al. (2023) contemporaneous to our work. We use the same pretrained model and test sets from our experiments in Section 4.1.

| Dataset    | Duplication Count | Sharded p (ours) | Percent Contaminated (Min-K%-Prob) |
|------------|-------------------|------------------|------------------------------------|
| BoolQ     | 1            | 0.156          | 3%                             |
| HellaSwag | 1            | 0.478          | 2%                            |
| MNLI     | 10            | 1.96e-11           | 100%                           |
| MMLU-Pro-Law | 50        |  1e-38           | 90%                         |
| MMLU-HS-Psych | 100     |  1e-38           | 74% |

Our run of Min-k%-Prob follows the methodology outlined in the paper; we run the method on one hundred 512-token spans sampled from each benchmark, and tune the decision threshold on a validation set of five of our contaminated test sets, and five test sets not used in our data mixture (uncontaminated). The threshold is tuned for a false positive rate of 5% to allow for a meaningful comparison against our test. A value of k=20 is used as is recommended in the paper.

We find that our method matches or exceeds the performance of this state of the art heuristic method, while also providing statistical proof of contamination.

Output:
[
    {
        "question": "Although a more efficient sharded rank comparison test is proposed, the computational complexity is still considerable. For example, testing 49 files using 1000 permutations per shard can take 12 hours for LLaMA2.",
        "answer": "We'd like to address the concern regarding the computational complexity of our test. It's important to note that the test is a one-time process for any given model and dataset; once the p-values are computed, there is no need for recalculation. Our findings indicate that a number of permutations beyond 30-50 per shard offers diminishing returns, as shown in Figure 3 (right). Furthermore, the test's design allows for easy parallelization. Each shard permutation can be evaluated independently, enabling the use of inexpensive commodity hardware to run the test significantly faster.",
        "is_multimodal_related": true
    },
    {
        "question": "There is no comparison with other baseline methods.",
        "answer": "Finally, we appreciate your suggestion to include baseline comparisons. We provide a comparison against a contamination detection method called Min-K% Prob, a state of the art heuristic method for contamination detection in language models proposed contemporaneous to our work by Shi et. al. (2023). We find that our method matches or exceeds the performance of this state of the art heuristic method. Please see the table in the top-level comment for numbers. While no other existing work is comparable in the sense that it provides a statistical proof of contamination like ours, we provide a comparison against a state of the art heuristic method for contamination detection called Min-K% Prob, proposed by Shi et. al. (2023) contemporaneous to our work. We use the same pretrained model and test sets from our experiments in Section 4.1.\n\n| Dataset    | Duplication Count | Sharded p (ours) | Percent Contaminated (Min-K%-Prob) |\n|------------|-------------------|------------------|------------------------------------|\n| BoolQ     | 1            | 0.156          | 3%                             |\n| HellaSwag | 1            | 0.478          | 2%                            |\n| MNLI     | 10            | 1.96e-11           | 100%                           |\n| MMLU-Pro-Law | 50        |  1e-38           | 90%                         |\n| MMLU-HS-Psych | 100     |  1e-38           | 74% |\n\nOur run of Min-k%-Prob follows the methodology outlined in the paper; we run the method on one hundred 512-token spans sampled from each benchmark, and tune the decision threshold on a validation set of five of our contaminated test sets, and five test sets not used in our data mixture (uncontaminated). The threshold is tuned for a false positive rate of 5% to allow for a meaningful comparison against our test. A value of k=20 is used as is recommended in the paper. We find that our method matches or exceeds the performance of this state of the art heuristic method, while also providing statistical proof of contamination.",
        "is_multimodal_related": false
    },
    {
        "question": "The method relies on a strong assumption of data exchangeability, which may not hold in real-world datasets.",
        "answer": "Regarding the assumption of data exchangeability, this is a strictly weaker condition than the commonly held assumption of independent and identically distributed (I.I.D.) data in machine learning. Most datasets satisfy this assumption to some extent.",
        "is_multimodal_related": false
    },
    {
        "question": "If a dataset is not exchangeable, how effective is the method?",
        "answer": "We acknowledge the validity of our test hinges on data exchangeability. However, depending on the source of non-exchangeability, it is often the case that a dataset can be altered slightly so that our test is still valid. For example, a common source of non-exchangeability is the presence of ascending IDs (e.g. as in SQuAD and HumanEval). We can adjust the data-by either removing these IDs or permuting the examples while keeping IDs constant-to retain the test's applicability. This is discussed in more detail in the revised paper.",
        "is_multimodal_related": false
    }
]
\end{lstlisting}

\subsection{Conversion Prompt} \label{sec:prompt_conversion}
\begin{lstlisting}[style=prompt]
You are an advanced assistant trained for academic research purposes. Your task is to process all review-rebuttal pairs into a structured Question-Answer (QA) format. For every input pair, follow these instructions:

Input Structure:
You will process all review-rebuttal pairs, where each is provided in the following format:
Review: A statement or query from a reviewer providing feedback or posing a question about the submission.
Rebuttal: The corresponding author response addressing the feedback.

Processing Instructions:
For each review-rebuttal pair, follow the steps below in strict sequence:
1. Extract the Question (Q):
Reformulate the reviewer feedback into a clear, precise, and standalone question. Ensure the question:
Includes all necessary context from both the review and rebuttal (e.g., clarify vague references such as "this figure" or "the results").
Is phrased in neutral and objective language, avoiding subjective or opinionated terms.
2. Extract the Answer (A):
Reformulate the author's rebuttal into a concise, objective, and standalone answer. Ensure the answer:
Directly addresses the reformulated question.
Is based strictly on the rebuttal content. Avoid additional interpretations, subjective language, or opinions.
3. Classify the Question:
Classify the question into a precise subcategory based on its intent using the schema below (see categories below).

Categories:
1. Concept Understanding [What]: Clarifies or explains key concepts, terminology, theoretical viewpoints, or information conveyed in figures, tables, or formulas.
2. Methods
    2.1. Method Disambiguation [What]: Clarifies methodological details to resolve misunderstandings or ambiguities, ensuring an accurate grasp of proposed approaches.
    2.2. Method Mechanics [How]: Questions about the implementation or function of methodological workflow or components, such as the effect of specific modules in models.
    2.3. Motivation Analysis [Why]: Examines the rationale, principles, or intentions underlying a proposed method or decision.
    2.4. Method Comparison : Compares the proposed approach with baseline methods, analyzing similarities, differences, or performance to highlight novelty.
3. Experiments
    3.1. Experimental Exposition [What]: Describes experimental outcomes, infers how modifications or variations could impact results or conclusions, and addresses reasoning tasks such as calculation, counting, or comparative analysis.
    3.2. Experimental Setup [How]: About the design, configuration, and execution of experiments.
    3.3. Experimental Analysis [Why]: Studies the reasons of specific experimental outcomes, links them to the proposed approach, and assesses their generalizability and potential impact.
4. Claim Verification : Binary classification tasks that assess the correctness of claims, hypotheses, or experimental conclusions.

Output Format: Provide the processed data for each review-rebuttal pair in the following JSON format:
[
    {
        "review": "Original reviewer feedback",
        "rebuttal": "Original author rebuttal",
        "Q": "Generated question",
        "A": "Generated answer",
        "Category": "Selected subcategory"
    },
    {
        "review": "Original reviewer feedback",
        "rebuttal": "Original author rebuttal",
        "Q": "Generated question",
        "A": "Generated answer",
        "Category": "Selected subcategory"
    },
    ...
]
\end{lstlisting}

\subsection{Reasoning Prompt} \label{sec:prompt_reasoning}

Open-ended QA: 
\begin{lstlisting}[style=prompt]
You are an expert academic assistant. Your task is to carefully read and analyze the provided complete research paper, and then answer the following question solely based on its content, arguments, and data, without using any external information or assumptions.
Response Requirements:
1. The answer must be professional, precise, concise, and clearly presented.
2. All statements in your answer must be exclusively derived from the paper's content and directly relevant to the question, avoiding any information or claims not supported by the paper.
3. The total length of your response must not exceed 3000 characters (including spaces).

Question:
{question}

Paper:
{content}
\end{lstlisting}

Claim verification: 
\begin{lstlisting}[style=prompt]
You are an academic judgment specialist assigned to classify the following statement as strictly 'True' or 'False' based exclusively on the content of the provided research paper. Carefully read and analyze the entire paper. Use only evidence directly from the text, and not incorporate external knowledge, assumptions, or subjective reasoning.

Output Requirements:
- Respond SOLELY with 'True' or 'False'
- No explanations, disclaimers, or supplementary text

Statement:
{question}

Paper:
{content}
\end{lstlisting}

\subsection{Evaluation Prompt} \label{sec:prompt_evaluation}

Message provided to the LLM during evaluation: 
\begin{lstlisting}[style=prompt]
messages = [
    {"role": "system", "content": sys_prompt},
    {"role": "user", "content": Conciseness/Correctness/Completeness.format(title=title, abstract=abstract, question=question, reference_answer=reference_answer, predicted_answer=predicted_answer)},
]
\end{lstlisting}

System prompt: 
\begin{lstlisting}[style=prompt]
Evaluate and rate the quality of the following predicted answer to an academic question according to the evaluation characteristics given in the system prompt.

<paper-title>{title}</paper-title>

<paper-abstract>{abstract}</paper-abstract>

<question>{question}</question>

<reference-answer>{reference_answer}</reference-answer>

<predicted-answer>{predicted_answer}</predicted-answer>
\end{lstlisting}

Conciseness: 
\begin{lstlisting}[style=prompt]
<Context>
Academic question answering is the process of thoroughly reading and analyzing a scientific paper in order to generate answers to specific questions based solely on the paper's content, arguments, and data. Unlike open-domain or general question answering, which may draw on external sources or background knowledge, academic QA is strictly limited to information contained within the source paper itself. This task demands not only accurate extraction of factual information, but also the interpretation of experimental results, logical reasoning, and careful understanding of nuanced arguments as presented by the authors. Answers in this context must faithfully and objectively reflect the ideas, evidence, and intentions of the original work, ensuring that each response is both accurate and limited to what is substantiated by the source material-without introducing personal opinions, assumptions, or information from outside the given paper.
</Context>

<Role>
You are an expert academic answer evaluator.
</Role>

<Task-Description>
The task is to evaluate the quality of a predicted answer to a given academic question. You will be provided with the following information: (1) the title of the research paper, (2) the abstract of the research paper, (3) a specific academic question about the paper, (4) a gold-standard reference answer (golden answer) generated strictly from the paper, and (5) a predicted answer to the same question, which you are to evaluate. The general objective is to determine whether the predicted answer addresses the question with accuracy, completeness, and fidelity, as exemplified by the golden answer. Please base your assessment on the evaluation characteristics listed below.
</Task-Description>

<Evaluation-Characteristics>
1. Conciseness: Evaluate whether the predicted answer is brief and to the point, avoiding unnecessary repetition or irrelevant information. The answer should deliver key content clearly, without excessive length or verbosity.
</Evaluation-Characteristics>

<Rating-Scale>
For each evaluation characteristic, assign a quality score between 0.00 (very bad) and 5.00 (very good), using decimal values precise to two decimal places (e.g., 3.73) for fine-grained assessment. Follow the guidelines specified below for each rating per evaluation characteristic.

1. Conciseness
0.00-1.00 (Very bad): The predicted answer is verbose or contains substantial irrelevant/redundant information, making it unclear or unfocused.
1.01-2.00 (Bad): The predicted answer includes some redundancy or unnecessary details, affecting clarity.
2.01-3.00 (Moderate): The predicted answer is generally clear but could benefit from further condensation to remove several minor redundancies.
3.01-4.00 (Good): The predicted answer is concise, with only minimal unnecessary information.
4.01-5.00 (Very good): The predicted answer is exceptionally concise, presenting essential information directly and clearly with no redundancy.
</Rating-Scale>

<Response-Format>
For each characteristic, rate the quality with a decimal score between 0.00 (very bad) and 5.00 (very good), precise to two decimal places (e.g., 4.21). Provide a short rationale for each rating. 
Return your response in JSON format: {characteristic : {"rating": "", "rationale": ""}}

<Example-Response>
{
  "Conciseness": {
    "rating": "4.15",
    "rationale": "The answer is generally concise and focused, with only minimal redundant information."
  }
}
</Example-Response>
</Response-Format>

<Note>
Base your evaluation solely on the paper title, abstract, question, golden answer, and predicted answer provided. Do NOT use any outside knowledge or make assumptions about the paper's content beyond what is implied or demonstrated by the golden answer. Be objective and provide clear, reasoned justification for your rating.
</Note>
\end{lstlisting}

Correctness:
\begin{lstlisting}[style=prompt]
<Context>
Academic question answering is the process of thoroughly reading and analyzing a scientific paper in order to generate answers to specific questions based solely on the paper's content, arguments, and data. Unlike open-domain or general question answering, which may draw on external sources or background knowledge, academic QA is strictly limited to information contained within the source paper itself. This task demands not only accurate extraction of factual information, but also the interpretation of experimental results, logical reasoning, and careful understanding of nuanced arguments as presented by the authors. Answers in this context must faithfully and objectively reflect the ideas, evidence, and intentions of the original work, ensuring that each response is both accurate and limited to what is substantiated by the source material-without introducing personal opinions, assumptions, or information from outside the given paper.
</Context>

<Role>
You are an expert academic answer evaluator.
</Role>

<Task-Description>
The task is to evaluate the quality of a predicted answer to a given academic question. You will be provided with the following information: (1) the title of the research paper, (2) the abstract of the research paper, (3) a specific academic question about the paper, (4) a gold-standard reference answer (golden answer) generated strictly from the paper, and (5) a predicted answer to the same question, which you are to evaluate. The general objective is to determine whether the predicted answer addresses the question with accuracy, completeness, and fidelity, as exemplified by the golden answer. Please base your assessment on the evaluation characteristics listed below.
</Task-Description>

<Evaluation-Characteristics>
1. Correctness: Assess the proportion of content from the reference answer that is accurately reflected in the predicted answer. This is analogous to precision-focus on the accuracy and fidelity of included information, ensuring no distortions or misrepresentations.
</Evaluation-Characteristics>

<Rating-Scale>
For each evaluation characteristic, assign a quality score between 0.00 (very bad) and 5.00 (very good), using decimal values precise to two decimal places (e.g., 3.73) for fine-grained assessment. Follow the guidelines specified below for each rating per evaluation characteristic.

1. Correctness
0.00-1.00 (Very bad): The predicted answer consistently misrepresents or distorts the content of the reference answer, with substantial factual errors.
1.01-2.00 (Bad): The predicted answer contains multiple inaccuracies or significant misinterpretations relative to the reference answer.
2.01-3.00 (Moderate): The predicted answer accurately includes some content from the reference answer but may also have minor misstatements or factual inaccuracies.
3.01-4.00 (Good): Most content from the reference answer is accurately represented in the predicted answer, with only rare errors.
4.01-5.00 (Very good): Virtually all content from the reference answer present in the predicted answer is accurate and faithful, with no factual errors or distortions.
</Rating-Scale>

<Response-Format>
For each characteristic, rate the quality with a decimal score between 0.00 (very bad) and 5.00 (very good), precise to two decimal places (e.g., 4.21). Provide a short rationale for each rating. 
Return your response in JSON format: {characteristic : {"rating": "", "rationale": ""}}

<Example-Response>
{
  "Correctness": {
    "rating": "4.03",
    "rationale": "Most of the information in the answer accurately reflects the reference answer, with only minor factual inaccuracies."
  }
}
</Example-Response>
</Response-Format>

<Note>
Base your evaluation solely on the paper title, abstract, question, golden answer, and predicted answer provided. Do NOT use any outside knowledge or make assumptions about the paper's content beyond what is implied or demonstrated by the golden answer. Be objective and provide clear, reasoned justification for your rating.
</Note>
\end{lstlisting}

Completeness:
\begin{lstlisting}[style=prompt]
<Context>
Academic question answering is the process of thoroughly reading and analyzing a scientific paper in order to generate answers to specific questions based solely on the paper's content, arguments, and data. Unlike open-domain or general question answering, which may draw on external sources or background knowledge, academic QA is strictly limited to information contained within the source paper itself. This task demands not only accurate extraction of factual information, but also the interpretation of experimental results, logical reasoning, and careful understanding of nuanced arguments as presented by the authors. Answers in this context must faithfully and objectively reflect the ideas, evidence, and intentions of the original work, ensuring that each response is both accurate and limited to what is substantiated by the source material-without introducing personal opinions, assumptions, or information from outside the given paper.
</Context>

<Role>
You are an expert academic answer evaluator.
</Role>

<Task-Description>
The task is to evaluate the quality of a predicted answer to a given academic question. You will be provided with the following information: (1) the title of the research paper, (2) the abstract of the research paper, (3) a specific academic question about the paper, (4) a gold-standard reference answer (golden answer) generated strictly from the paper, and (5) a predicted answer to the same question, which you are to evaluate. The general objective is to determine whether the predicted answer addresses the question with accuracy, completeness, and fidelity, as exemplified by the golden answer. Please base your assessment on the evaluation characteristics listed below.
</Task-Description>

<Evaluation-Characteristics>
1. Completeness: Assess the proportion of information in the predicted answer that overlaps with the reference answer. This is analogous to recall-consider whether the predicted answer adequately covers all major points and details provided by the reference answer, and does not omit essential content.
</Evaluation-Characteristics>

<Rating-Scale>
For each evaluation characteristic, assign a quality score between 0.00 (very bad) and 5.00 (very good), using decimal values precise to two decimal places (e.g., 3.73) for fine-grained assessment. Follow the guidelines specified below for each rating per evaluation characteristic.

1. Completeness
0.00-1.00 (Very bad): The predicted answer fails to include most of the key content from the reference answer, omitting essential points or details.
1.01-2.00 (Bad): The predicted answer is missing several important aspects found in the reference answer.
2.01-3.00 (Moderate): The predicted answer includes a moderate portion of the relevant content from the reference answer but lacks full coverage.
3.01-4.00 (Good): Most relevant content from the reference answer is present, with only minor omissions.
4.01-5.00 (Very good): The predicted answer comprehensively incorporates all major information from the reference answer, leaving out nothing significant.
</Rating-Scale>

<Response-Format>
For each characteristic, rate the quality with a decimal score between 0.00 (very bad) and 5.00 (very good), precise to two decimal places (e.g., 4.21). Provide a short rationale for each rating. 
Return your response in JSON format: {characteristic : {"rating": "", "rationale": ""}}

<Example-Response>
{
  "Completeness": {
    "rating": "3.52",
    "rationale": "The answer covers most of the key points from the reference answer, but omits a few minor details."
  }
}
</Example-Response>
</Response-Format>

<Note>
Base your evaluation solely on the paper title, abstract, question, golden answer, and predicted answer provided. Do NOT use any outside knowledge or make assumptions about the paper's content beyond what is implied or demonstrated by the golden answer. Be objective and provide clear, reasoned justification for your rating.
</Note>
\end{lstlisting}

%% file: 0-paper-qa.bbl
\begin{thebibliography}{43}
\providecommand{\natexlab}[1]{#1}

\bibitem[{Anthropic(2025)}]{claude4}
Anthropic. 2025.
\newblock Introducing claude 4.
\newblock \url{https://www.anthropic.com/news/claude-4}.
\newblock Accessed: 2025-05-23.

\bibitem[{Asai et~al.(2024)Asai, He, Shao, Shi, Singh, Chang, Lo, Soldaini, Feldman, D'arcy et~al.}]{asai2024openscholar}
Akari Asai, Jacqueline He, Rulin Shao, Weijia Shi, Amanpreet Singh, Joseph~Chee Chang, Kyle Lo, Luca Soldaini, Sergey Feldman, Mike D'arcy, and 1 others. 2024.
\newblock Openscholar: Synthesizing scientific literature with retrieval-augmented lms.
\newblock \emph{arXiv preprint arXiv:2411.14199}.

\bibitem[{Bai et~al.(2023)Bai, Bai, Chu, Cui, Dang, Deng, Fan, Ge, Han, Huang, Hui, Ji, Li, Lin, Lin, Liu, Liu, Lu, Lu, Ma, Men, Ren, Ren, Tan, Tan, Tu, Wang, Wang, Wang, Wu, Xu, Xu, Yang, Yang, Yang, Yang, Yao, Yu, Yuan, Yuan, Zhang, Zhang, Zhang, Zhang, Zhou, Zhou, Zhou, and Zhu}]{bai2023qwentechnicalreport}
Jinze Bai, Shuai Bai, Yunfei Chu, Zeyu Cui, Kai Dang, Xiaodong Deng, Yang Fan, Wenbin Ge, Yu~Han, Fei Huang, Binyuan Hui, Luo Ji, Mei Li, Junyang Lin, Runji Lin, Dayiheng Liu, Gao Liu, Chengqiang Lu, Keming Lu, and 29 others. 2023.
\newblock \href {https://arxiv.org/abs/2309.16609} {Qwen technical report}.
\newblock \emph{Preprint}, arXiv:2309.16609.

\bibitem[{Baumg{\"a}rtner et~al.(2025)Baumg{\"a}rtner, Briscoe, and Gurevych}]{baumgartner2025peerqa}
Tim Baumg{\"a}rtner, Ted Briscoe, and Iryna Gurevych. 2025.
\newblock Peerqa: A scientific question answering dataset from peer reviews.
\newblock \emph{arXiv preprint arXiv:2502.13668}.

\bibitem[{Booth et~al.(2009)Booth, Colomb, and Williams}]{booth2009craft}
Wayne~C Booth, Gregory~G Colomb, and Joseph~M Williams. 2009.
\newblock \emph{The craft of research}.
\newblock University of Chicago press.

\bibitem[{Chen et~al.(2025)Chen, Zhang, and Tang}]{chen2025small}
Yelin Chen, Fanjin Zhang, and Jie Tang. 2025.
\newblock Small language model makes an effective long text extractor.
\newblock In \emph{Proceedings of the AAAI Conference on Artificial Intelligence}, volume~39, pages 23623--23631.

\bibitem[{Comanici et~al.(2025)Comanici, Bieber, Schaekermann, Pasupat, Sachdeva, Dhillon, Blistein, Ram, Zhang, Rosen et~al.}]{comanici2025gemini}
Gheorghe Comanici, Eric Bieber, Mike Schaekermann, Ice Pasupat, Noveen Sachdeva, Inderjit Dhillon, Marcel Blistein, Ori Ram, Dan Zhang, Evan Rosen, and 1 others. 2025.
\newblock Gemini 2.5: Pushing the frontier with advanced reasoning, multimodality, long context, and next generation agentic capabilities.
\newblock \emph{arXiv preprint arXiv:2507.06261}.

\bibitem[{Dasigi et~al.(2021)Dasigi, Lo, Beltagy, Cohan, Smith, and Gardner}]{dasigi2021dataset}
Pradeep Dasigi, Kyle Lo, Iz~Beltagy, Arman Cohan, Noah~A Smith, and Matt Gardner. 2021.
\newblock A dataset of information-seeking questions and answers anchored in research papers.
\newblock In \emph{Proceedings of the 2021 Conference of the North American Chapter of the Association for Computational Linguistics: Human Language Technologies}, pages 4599--4610.

\bibitem[{Deng et~al.(2024)Deng, Yuan, Bu, Wang, Li, Xu, Li, Gao, Song, Zheng et~al.}]{deng2024longdocurl}
Chao Deng, Jiale Yuan, Pi~Bu, Peijie Wang, Zhong-Zhi Li, Jian Xu, Xiao-Hui Li, Yuan Gao, Jun Song, Bo~Zheng, and 1 others. 2024.
\newblock Longdocurl: a comprehensive multimodal long document benchmark integrating understanding, reasoning, and locating.
\newblock \emph{arXiv preprint arXiv:2412.18424}.

\bibitem[{Desmond et~al.(2025)Desmond, Ashktorab, Geyer, Daly, Cooper, Pan, Nair, Wagner, and Pedapati}]{desmond2025evalassist}
Michael Desmond, Zahra Ashktorab, Werner Geyer, Elizabeth~M Daly, Martin~Santillan Cooper, Qian Pan, Rahul Nair, Nico Wagner, and Tejaswini Pedapati. 2025.
\newblock Evalassist: Llm-as-a-judge simplified.
\newblock In \emph{Proceedings of the AAAI Conference on Artificial Intelligence}, volume~39, pages 29637--29639.

\bibitem[{D'Souza et~al.(2025)D'Souza, Giglou, and M{\"u}nch}]{d2025yescieval}
Jennifer D'Souza, Hamed~Babaei Giglou, and Quentin M{\"u}nch. 2025.
\newblock Yescieval: Robust llm-as-a-judge for scientific question answering.
\newblock \emph{arXiv preprint arXiv:2505.14279}.

\bibitem[{Duan et~al.(2025)Duan, Chen, Hu, Wang, Ye, Shi, Lu, Hou, Lu, Li et~al.}]{duan2025docopilot}
Yuchen Duan, Zhe Chen, Yusong Hu, Weiyun Wang, Shenglong Ye, Botian Shi, Lewei Lu, Qibin Hou, Tong Lu, Hongsheng Li, and 1 others. 2025.
\newblock Docopilot: Improving multimodal models for document-level understanding.
\newblock In \emph{Proceedings of the Computer Vision and Pattern Recognition Conference}, pages 4026--4037.

\bibitem[{Edge et~al.(2025)Edge, Trinh, Cheng, Bradley, Chao, Mody, Truitt, Metropolitansky, Ness, and Larson}]{edge2025localglobalgraphrag}
Darren Edge, Ha~Trinh, Newman Cheng, Joshua Bradley, Alex Chao, Apurva Mody, Steven Truitt, Dasha Metropolitansky, Robert~Osazuwa Ness, and Jonathan Larson. 2025.
\newblock \href {https://arxiv.org/abs/2404.16130} {From local to global: A graph rag approach to query-focused summarization}.
\newblock \emph{Preprint}, arXiv:2404.16130.

\bibitem[{GLM et~al.(2024)GLM, :, Zeng, Xu, Wang, Zhang, Yin, Zhang, Rojas, Feng, Zhao, Lai, Yu, Wang, Sun, Zhang, Cheng, Gui, Tang, Zhang, Sun, Li, Zhao, Wu, Zhong, Liu, Huang, Zhang, Zheng, Lu, Duan, Zhang, Cao, Yang, Tam, Zhao, Liu, Xia, Zhang, Gu, Lv, Liu, Liu, Yang, Song, Zhang, An, Xu, Niu, Yang, Li, Bai, Dong, Qi, Wang, Yang, Du, Hou, and Wang}]{glm2024chatglmfamilylargelanguage}
Team GLM, :, Aohan Zeng, Bin Xu, Bowen Wang, Chenhui Zhang, Da~Yin, Dan Zhang, Diego Rojas, Guanyu Feng, Hanlin Zhao, Hanyu Lai, Hao Yu, Hongning Wang, Jiadai Sun, Jiajie Zhang, Jiale Cheng, Jiayi Gui, Jie Tang, and 40 others. 2024.
\newblock \href {https://arxiv.org/abs/2406.12793} {Chatglm: A family of large language models from glm-130b to glm-4 all tools}.
\newblock \emph{Preprint}, arXiv:2406.12793.

\bibitem[{Gottweis et~al.(2025)Gottweis, Weng, Daryin, Tu, Palepu, Sirkovic, Myaskovsky, Weissenberger, Rong, Tanno et~al.}]{gottweis2025towards}
Juraj Gottweis, Wei-Hung Weng, Alexander Daryin, Tao Tu, Anil Palepu, Petar Sirkovic, Artiom Myaskovsky, Felix Weissenberger, Keran Rong, Ryutaro Tanno, and 1 others. 2025.
\newblock Towards an ai co-scientist.
\newblock \emph{arXiv preprint arXiv:2502.18864}.

\bibitem[{Guti{\'e}rrez et~al.(2025)Guti{\'e}rrez, Shu, Qi, Zhou, and Su}]{gutierrez2025rag}
Bernal~Jim{\'e}nez Guti{\'e}rrez, Yiheng Shu, Weijian Qi, Sizhe Zhou, and Yu~Su. 2025.
\newblock From rag to memory: Non-parametric continual learning for large language models.
\newblock \emph{arXiv preprint arXiv:2502.14802}.

\bibitem[{Hu et~al.(2024)Hu, Xu, Zhang, Ye, Yan, Zhang, Jin, Huang, and Zhou}]{hu2024mplugdocowl2highresolutioncompressingocrfree}
Anwen Hu, Haiyang Xu, Liang Zhang, Jiabo Ye, Ming Yan, Ji~Zhang, Qin Jin, Fei Huang, and Jingren Zhou. 2024.
\newblock \href {https://arxiv.org/abs/2409.03420} {mplug-docowl2: High-resolution compressing for ocr-free multi-page document understanding}.
\newblock \emph{Preprint}, arXiv:2409.03420.

\bibitem[{Jin et~al.(2025)Jin, Zhu, Dou, Dong, Yang, Zhang, Zhao, Yang, and Wen}]{jin2025flashrag}
Jiajie Jin, Yutao Zhu, Zhicheng Dou, Guanting Dong, Xinyu Yang, Chenghao Zhang, Tong Zhao, Zhao Yang, and Ji-Rong Wen. 2025.
\newblock Flashrag: A modular toolkit for efficient retrieval-augmented generation research.
\newblock In \emph{Companion Proceedings of the ACM on Web Conference 2025}, pages 737--740.

\bibitem[{Krithara et~al.(2023)Krithara, Nentidis, Bougiatiotis, and Paliouras}]{krithara2023bioasq}
Anastasia Krithara, Anastasios Nentidis, Konstantinos Bougiatiotis, and Georgios Paliouras. 2023.
\newblock Bioasq-qa: A manually curated corpus for biomedical question answering.
\newblock \emph{Scientific data}, 10(1):170.

\bibitem[{Lee et~al.(2023)Lee, Lee, Park, Hwang, Kim, Lee, and Lee}]{lee2023qasa}
Yoonjoo Lee, Kyungjae Lee, Sunghyun Park, Dasol Hwang, Jaehyeon Kim, Hong-in Lee, and Moontae Lee. 2023.
\newblock Qasa: advanced question answering on scientific articles.
\newblock In \emph{International Conference on Machine Learning}, pages 19036--19052. PMLR.

\bibitem[{Leon(2025)}]{leon2025gpt}
Maikel Leon. 2025.
\newblock Gpt-5 and open-weight large language models: Advances in reasoning, transparency, and control.
\newblock \emph{Information Systems}, page 102620.

\bibitem[{Li et~al.(2024{\natexlab{a}})Li, Wang, Xu, Wang, Feng, Kong, and Liu}]{li2024multimodal}
Lei Li, Yuqi Wang, Runxin Xu, Peiyi Wang, Xiachong Feng, Lingpeng Kong, and Qi~Liu. 2024{\natexlab{a}}.
\newblock Multimodal arxiv: A dataset for improving scientific comprehension of large vision-language models.
\newblock \emph{arXiv preprint arXiv:2403.00231}.

\bibitem[{Li et~al.(2024{\natexlab{b}})Li, Yang, Liu, Ma, Zhang, Yang, Sun, Liu, and Bai}]{li2024monkeyimageresolutiontext}
Zhang Li, Biao Yang, Qiang Liu, Zhiyin Ma, Shuo Zhang, Jingxu Yang, Yabo Sun, Yuliang Liu, and Xiang Bai. 2024{\natexlab{b}}.
\newblock \href {https://arxiv.org/abs/2311.06607} {Monkey: Image resolution and text label are important things for large multi-modal models}.
\newblock \emph{Preprint}, arXiv:2311.06607.

\bibitem[{Liu et~al.(2025)Liu, Mei, Lin, Xue, Wang, Xu, Wu, Zhang, Lin, Dong et~al.}]{liu2025deepseek}
Aixin Liu, Aoxue Mei, Bangcai Lin, Bing Xue, Bingxuan Wang, Bingzheng Xu, Bochao Wu, Bowei Zhang, Chaofan Lin, Chen Dong, and 1 others. 2025.
\newblock Deepseek-v3. 2: Pushing the frontier of open large language models.
\newblock \emph{arXiv preprint arXiv:2512.02556}.

\bibitem[{Pramanick et~al.(2024)Pramanick, Chellappa, and Venugopalan}]{pramanick2024spiqa}
Shraman Pramanick, Rama Chellappa, and Subhashini Venugopalan. 2024.
\newblock Spiqa: A dataset for multimodal question answering on scientific papers.
\newblock \emph{Advances in Neural Information Processing Systems}, 37:118807--118833.

\bibitem[{Qian et~al.(2025)Qian, Liu, Zhang, Mao, Lian, Dou, and Huang}]{qian2025memorag}
Hongjin Qian, Zheng Liu, Peitian Zhang, Kelong Mao, Defu Lian, Zhicheng Dou, and Tiejun Huang. 2025.
\newblock Memorag: Boosting long context processing with global memory-enhanced retrieval augmentation.
\newblock In \emph{Proceedings of the ACM on Web Conference 2025}, pages 2366--2377.

\bibitem[{Sarthi et~al.(2024)Sarthi, Abdullah, Tuli, Khanna, Goldie, and Manning}]{sarthi2024raptor}
Parth Sarthi, Salman Abdullah, Aditi Tuli, Shubh Khanna, Anna Goldie, and Christopher~D Manning. 2024.
\newblock Raptor: Recursive abstractive processing for tree-organized retrieval.
\newblock In \emph{The Twelfth International Conference on Learning Representations}.

\bibitem[{Schmidgall et~al.(2025)Schmidgall, Su, Wang, Sun, Wu, Yu, Liu, Moor, Liu, and Barsoum}]{schmidgall2025agentlaboratoryusingllm}
Samuel Schmidgall, Yusheng Su, Ze~Wang, Ximeng Sun, Jialian Wu, Xiaodong Yu, Jiang Liu, Michael Moor, Zicheng Liu, and Emad Barsoum. 2025.
\newblock \href {https://arxiv.org/abs/2501.04227} {Agent laboratory: Using llm agents as research assistants}.
\newblock \emph{Preprint}, arXiv:2501.04227.

\bibitem[{Sollaci and Pereira(2004)}]{sollaci2004introduction}
Luciana~B Sollaci and Mauricio~G Pereira. 2004.
\newblock The introduction, methods, results, and discussion (imrad) structure: a fifty-year survey.
\newblock \emph{Journal of the medical library association}, 92(3):364.

\bibitem[{Tanaka et~al.(2025)Tanaka, Iki, Hasegawa, Nishida, Saito, and Suzuki}]{tanaka2025vdocrag}
Ryota Tanaka, Taichi Iki, Taku Hasegawa, Kyosuke Nishida, Kuniko Saito, and Jun Suzuki. 2025.
\newblock Vdocrag: Retrieval-augmented generation over visually-rich documents.
\newblock In \emph{Proceedings of the Computer Vision and Pattern Recognition Conference}, pages 24827--24837.

\bibitem[{Tang et~al.(2008)Tang, Zhang, Yao, Li, Zhang, and Su}]{tang2008arnetminer}
Jie Tang, Jing Zhang, Limin Yao, Juanzi Li, Li~Zhang, and Zhong Su. 2008.
\newblock Arnetminer: extraction and mining of academic social networks.
\newblock In \emph{Proceedings of the 14th ACM SIGKDD international conference on Knowledge discovery and data mining}, pages 990--998.

\bibitem[{Team et~al.(2025{\natexlab{a}})Team, Zeng, Lv, Zheng, Hou, Chen, Xie, Wang, Yin, Zeng, Zhang, Wang, Zhong, Liu, Lu, Cao, Zhang, Huang, Wei, Cheng, An, Niu, Wen, Bai, Du, Wang, Zhu, Zhang, Wen, Wu, Xu, Huang, Zhao, Cai, Yu, Li, Ge, Huang, Zhang, Xu, Zhu, Li, Yin, Lin, Yang, Jiang, Ai, Zhu, Wang, Pan, Wang, Sun, Li, Li, Hu, Zhang, Peng, Tai, Zhang, Wang, Yang, Liu, Zhao, Liu, Yan, Liu, Chen, Li, Zhao, Ren, Jiao, Zhao, Yan, Wang, Gui, Zhao, Liu, Li, Li, Lu, Wang, Yuan, Li, Du, Du, Liu, Zhi, Gao, Wang, Yang, Xu, Fan, Wu, Ding, Wang, Zhang, Li, Xu, Zhao, Zhai, Du, Dong, Lei, Tu, Yang, Lu, Li, Li, Shuang-Li, Yang, Yi, Yu, Tian, Wang, Yu, Tam, Liang, Liu, Wang, Jia, Gu, Ling, Wang, Fan, Pan, Zhang, Zhang, Fu, Zhang, Xu, Wu, Lu, Wang, Zhou, Pan, Zhang, Wang, Li, Su, Geng, Zhu, Yang, Li, Wu, Li, Liu, Wang, Li, Zhang, Liu, Yang, Zhou, Qiao, Feng, Liu, Zhang, Wang, Yao, Wang, Liu, Chai, Li, Zhao, Chen, Zhai, Xu, Huang, Wang, Li, Dong, and Tang}]{5team2025glm45agenticreasoningcoding}
GLM Team, Aohan Zeng, Xin Lv, Qinkai Zheng, Zhenyu Hou, Bin Chen, Chengxing Xie, Cunxiang Wang, Da~Yin, Hao Zeng, Jiajie Zhang, Kedong Wang, Lucen Zhong, Mingdao Liu, Rui Lu, Shulin Cao, Xiaohan Zhang, Xuancheng Huang, Yao Wei, and 152 others. 2025{\natexlab{a}}.
\newblock \href {https://arxiv.org/abs/2508.06471} {Glm-4.5: Agentic, reasoning, and coding (arc) foundation models}.
\newblock \emph{Preprint}, arXiv:2508.06471.

\bibitem[{Team et~al.(2025{\natexlab{b}})Team, Hong, Yu, Gu, Wang, Gan, Tang, Cheng, Qi, Ji, Pan, Duan, Wang, Wang, Cheng, He, Su, Yang, Pan, Zeng, Wang, Chen, Shi, Pang, Zhang, Yin, Yang, Chen, Xu, Zhu, Chen, Chen, Chen, Lin, Wang, Chen, Lei, Gong, Pan, Liu, Xu, Zhang, Zheng, Yang, Zhong, Huang, Zhao, Xue, Tu, Meng, Zhang, Luo, Hao, Tong, Li, Jia, Liu, Zhang, Lyu, Fan, Huang, Wang, Xue, Wang, Wang, An, Du, Shi, Huang, Niu, Wang, Yue, Li, Zhang, Wang, Wang, Zhang, Xue, Hou, Du, Wang, Zhang, Liu, Xu, Li, Huang, Dong, and Tang}]{vteam2025glm45vglm41vthinkingversatilemultimodal}
V~Team, Wenyi Hong, Wenmeng Yu, Xiaotao Gu, Guo Wang, Guobing Gan, Haomiao Tang, Jiale Cheng, Ji~Qi, Junhui Ji, Lihang Pan, Shuaiqi Duan, Weihan Wang, Yan Wang, Yean Cheng, Zehai He, Zhe Su, Zhen Yang, Ziyang Pan, and 69 others. 2025{\natexlab{b}}.
\newblock \href {https://arxiv.org/abs/2507.01006} {Glm-4.5v and glm-4.1v-thinking: Towards versatile multimodal reasoning with scalable reinforcement learning}.
\newblock \emph{Preprint}, arXiv:2507.01006.

\bibitem[{Turner and Firth(2012)}]{turner2012bradley}
Heather Turner and David Firth. 2012.
\newblock Bradley-terry models in r: the bradleyterry2 package.
\newblock \emph{Journal of statistical software}, 48:1--21.

\bibitem[{Wadden et~al.(2020)Wadden, Lin, Lo, Wang, van Zuylen, Cohan, and Hajishirzi}]{wadden2020fact}
David Wadden, Shanchuan Lin, Kyle Lo, Lucy~Lu Wang, Madeleine van Zuylen, Arman Cohan, and Hannaneh Hajishirzi. 2020.
\newblock Fact or fiction: Verifying scientific claims.
\newblock In \emph{Proceedings of the 2020 Conference on Empirical Methods in Natural Language Processing (EMNLP)}, pages 7534--7550.

\bibitem[{Wang et~al.(2023)Wang, Raman, Sibue, Ma, Babkin, Kaur, Pei, Nourbakhsh, and Liu}]{wang2023docllm}
Dongsheng Wang, Natraj Raman, Mathieu Sibue, Zhiqiang Ma, Petr Babkin, Simerjot Kaur, Yulong Pei, Armineh Nourbakhsh, and Xiaomo Liu. 2023.
\newblock Docllm: A layout-aware generative language model for multimodal document understanding.
\newblock \emph{arXiv preprint arXiv:2401.00908}.

\bibitem[{Xia et~al.(2024)Xia, Mao, Yan, Zhou, Zhang, Peng, Pi, Fu, Wu, Ye et~al.}]{xia2024docgenome}
Renqiu Xia, Song Mao, Xiangchao Yan, Hongbin Zhou, Bo~Zhang, Haoyang Peng, Jiahao Pi, Daocheng Fu, Wenjie Wu, Hancheng Ye, and 1 others. 2024.
\newblock Docgenome: An open large-scale scientific document benchmark for training and testing multi-modal large language models.
\newblock \emph{arXiv preprint arXiv:2406.11633}.

\bibitem[{Yamada et~al.(2025)Yamada, Lange, Lu, Hu, Lu, Foerster, Clune, and Ha}]{aiscientist_v2}
Yutaro Yamada, Robert~Tjarko Lange, Cong Lu, Shengran Hu, Chris Lu, Jakob Foerster, Jeff Clune, and David Ha. 2025.
\newblock The ai scientist-v2: Workshop-level automated scientific discovery via agentic tree search.
\newblock \emph{arXiv preprint arXiv:2504.08066}.

\bibitem[{Yang et~al.(2025)Yang, Li, Yang, Zhang, Hui, Zheng, Yu, Gao, Huang, Lv, Zheng, Liu, Zhou, Huang, Hu, Ge, Wei, Lin, Tang, Yang, Tu, Zhang, Yang, Yang, Zhou, Zhou, Lin, Dang, Bao, Yang, Yu, Deng, Li, Xue, Li, Zhang, Wang, Zhu, Men, Gao, Liu, Luo, Li, Tang, Yin, Ren, Wang, Zhang, Ren, Fan, Su, Zhang, Zhang, Wan, Liu, Wang, Cui, Zhang, Zhou, and Qiu}]{yang2025qwen3technicalreport}
An~Yang, Anfeng Li, Baosong Yang, Beichen Zhang, Binyuan Hui, Bo~Zheng, Bowen Yu, Chang Gao, Chengen Huang, Chenxu Lv, Chujie Zheng, Dayiheng Liu, Fan Zhou, Fei Huang, Feng Hu, Hao Ge, Haoran Wei, Huan Lin, Jialong Tang, and 41 others. 2025.
\newblock \href {https://arxiv.org/abs/2505.09388} {Qwen3 technical report}.
\newblock \emph{Preprint}, arXiv:2505.09388.

\bibitem[{Yifei et~al.(2025)Yifei, Chang, Malaviya, and Yatskar}]{yifei2025researchqa}
Li~S Yifei, Allen Chang, Chaitanya Malaviya, and Mark Yatskar. 2025.
\newblock Researchqa: Evaluating scholarly question answering at scale across 75 fields with survey-mined questions and rubrics.
\newblock \emph{arXiv preprint arXiv:2509.00496}.

\bibitem[{Yu et~al.(2024)Yu, Tang, Xu, Cui, Ran, Yan, Liu, Wang, Han, Liu et~al.}]{yu2024visrag}
Shi Yu, Chaoyue Tang, Bokai Xu, Junbo Cui, Junhao Ran, Yukun Yan, Zhenghao Liu, Shuo Wang, Xu~Han, Zhiyuan Liu, and 1 others. 2024.
\newblock Visrag: Vision-based retrieval-augmented generation on multi-modality documents.
\newblock \emph{arXiv preprint arXiv:2410.10594}.

\bibitem[{Zhang et~al.(2019)Zhang, Liu, Tang, Dong, Yao, Zhang, Gu, Wang, Shao, Li et~al.}]{zhang2019oag}
Fanjin Zhang, Xiao Liu, Jie Tang, Yuxiao Dong, Peiran Yao, Jie Zhang, Xiaotao Gu, Yan Wang, Bin Shao, Rui Li, and 1 others. 2019.
\newblock Oag: Toward linking large-scale heterogeneous entity graphs.
\newblock In \emph{Proceedings of the 25th ACM SIGKDD International Conference on Knowledge Discovery and Data Mining}, pages 2585--2595.

\bibitem[{Zhang et~al.(2024)Zhang, Shi, Zhu, Chen, Cen, Yu, Chen, Wang, Zhao, Cheng et~al.}]{zhang2024oag}
Fanjin Zhang, Shijie Shi, Yifan Zhu, Bo~Chen, Yukuo Cen, Jifan Yu, Yelin Chen, Lulu Wang, Qingfei Zhao, Yuqing Cheng, and 1 others. 2024.
\newblock Oag-bench: a human-curated benchmark for academic graph mining.
\newblock In \emph{Proceedings of the 30th ACM SIGKDD Conference on Knowledge Discovery and Data Mining}, pages 6214--6225.

\end{thebibliography}
